\documentclass[twocolumn,letterpaper]{IEEEAerospaceCLS}

\usepackage[utf8]{inputenc}

\usepackage{amsmath, amssymb} %
\usepackage{multicol}

\usepackage{bm} %
\usepackage{siunitx} %

\usepackage{hyperref} %
\usepackage[noabbrev]{cleveref} %

\usepackage{multirow}

\usepackage{subfiles}
\usepackage{breqn}

\usepackage{ifthen}

\usepackage{graphicx}
\graphicspath{{images/}} %
\usepackage{caption}
\usepackage{subcaption}
\usepackage{tikz, tikz-3dplot}
\usetikzlibrary{calc, 3d}

\usepackage[colorinlistoftodos]{todonotes} %
\presetkeys{todonotes}{inline}{} %

\usepackage{biblatex}
\let\oldmathbb\mathbb
\renewcommand{\mathbb}[1]{{\oldmathbb{#1}}}
\let\oldmathbf\mathbf
\renewcommand{\mathbf}[1]{{\oldmathbf{#1}}}
\let\oldmathcal\mathcal
\renewcommand{\mathcal}[1]{{\oldmathcal{#1}}}

\newcommand{\decrement}[1]{%
\number\numexpr#1-1\relax%
}

\newcommand\sierpinskigasket[2][0]{
	\node (a) at ({1/sqrt(3)}, 0, {-1/3 * sqrt(2/3)}) {};
	\node (b) at ({-1/2/sqrt(3)}, {1/2}, {-1/3 * sqrt(2/3)}) {};
	\node (c) at ({-1/2/sqrt(3)}, {-1/2}, {-1/3 * sqrt(2/3)}) {};
	\node (d) at (0, 0, {2/3 * sqrt(2/3)}) {};
	
	\ifnum #2 = 0 \relax
		\fill[gray!40!] (a.center) -- (b.center) -- (d.center) -- cycle;
		\fill[gray!50!] (a.center) -- (c.center) -- (d.center) -- cycle;
	\else
	\begin{scope}[scale=.5]
		\begin{scope}[shift={({-1/2/sqrt(3)}, {1/2}, {-1/3 * sqrt(2/3)})}]
 			\sierpinskigasket{\numexpr#2-1}
		\end{scope}
		\begin{scope}[shift={({-1/2/sqrt(3)}, {-1/2}, {-1/3 * sqrt(2/3)})}]
 			\sierpinskigasket{\numexpr#2-1}
		\end{scope}
		\begin{scope}[shift={(0, 0, {2/3 * sqrt(2/3)})}]
		 	\sierpinskigasket{\numexpr#2-1}
		\end{scope}
		\begin{scope}[shift={({1/sqrt(3)}, 0, {-1/3 * sqrt(2/3)})}]
			\sierpinskigasket{\numexpr#2-1}
		\end{scope}
	\end{scope}
	\fi
	
	\ifnum #1 = 1 \relax
		\node (a) at ({1/2/sqrt(3)}, 0, {-1/6 * sqrt(2/3)}) {};
		\node (b) at ({-1/4/sqrt(3)}, {1/4}, {-1/6 * sqrt(2/3)}) {};
		\node (c) at ({-1/4/sqrt(3)}, {-1/4}, {-1/6 * sqrt(2/3)}) {};
		\node (d) at (0, 0, {2/6 * sqrt(2/3)}) {};

		\draw[->, thick] (0, 0, 0) node[above right] {$O$} -- (b.center) node[right] {$p^\decrement{#2}_2$};
		\draw[->, thick] (0, 0, 0) -- (c.center) node[left] {$p^\decrement{#2}_3$};
		\draw[->, thick] (0, 0, 0) -- (d.center) node[above right] {$p^\decrement{#2}_4$};
		\draw[->, thick] (0, 0, 0) -- (a.center) node[below] {$p^\decrement{#2}_1$};
	\fi
}

\makeatletter
\tikzoption{canvas is plane}[]{\@setOxy#1}
\def\@setOxy O(#1)x(#2)y(#3)%
  {\def\tikz@plane@origin{#1}%
   \def\tikz@plane@x{#2}%
   \def\tikz@plane@y{#3}%
   \tikz@canvas@is@plane
  }
\makeatother

\tikzset{
	pics/tetraprop/.style n args={2}{
	code = { %
	      \node (a) at ({1/sqrt(3)}, 0, 0) {};
	      \node (b) at ({-1/2/sqrt(3)}, { 1/2}, 0) {};
	      \node (c) at ({-1/2/sqrt(3)}, {-1/2}, 0) {};
	      \node (d) at (0, 0, {1 * sqrt(2/3)}) {};
	      
	      \coordinate (abc) at (barycentric cs:a=1,b=1,c=1);
	      \coordinate (abd) at (barycentric cs:a=1,b=1,d=1);
	      \coordinate (bcd) at (barycentric cs:b=1,c=1,d=1);
	      \coordinate (acd) at (barycentric cs:a=1,c=1,d=1);
	      
		\ifthenelse{#1 = 0}
		{
		}
		{
			\node (abdn) at (barycentric cs:abd=1,c=1.5) {};
			\draw [thick,orange,->,>=stealth] (abd) -- (abdn);
		}
		;
	      
	      \coordinate (abcu) at ($(abc)+(b)-(a)$);
	      \coordinate (abcv) at (barycentric cs:abc={1-sqrt(3)},c={sqrt(3)});
	      \coordinate (abdu) at ($(abd)+(b)-(a)$);
	      \coordinate (abdv) at (barycentric cs:abd={1-sqrt(3)},d={sqrt(3)});
	      \coordinate (bcdu) at ($(bcd)+(c)-(b)$);
	      \coordinate (bcdv) at (barycentric cs:bcd={1-sqrt(3)},d={sqrt(3)});
	      \coordinate (acdu) at ($(acd)+(c)-(a)$);
	      \coordinate (acdv) at (barycentric cs:acd={1-sqrt(3)},d={sqrt(3)});
	      
	      \draw[thick] (a.center) -- (b.center);
	      \draw (b.center) -- (c.center);
	      \draw (c.center) -- (a.center);
	      \draw[thick] (a.center) -- (d.center);
	      \draw[thick] (b.center) -- (d.center);
	      \draw (c.center) -- (d.center);
	      
	      \begin{scope}[canvas is plane={O(\pgfpointanchor{abc}{center})x(\pgfpointanchor{abcu}{center})y(\pgfpointanchor{abcv}{center})}]
	      \draw (abc) circle ({1 / 2 / sqrt(3)});
	      \end{scope}
	      	      
   	      \begin{scope}[canvas is plane={O(\pgfpointanchor{bcd}{center})x(\pgfpointanchor{bcdu}{center})y(\pgfpointanchor{bcdv}{center})}]
   	      \draw (bcd) circle ({1 / 2 / sqrt(3)});
   	      \end{scope}
   	      
   	      \begin{scope}[canvas is plane={O(\pgfpointanchor{acd}{center})x(\pgfpointanchor{acdu}{center})y(\pgfpointanchor{acdv}{center})}]
   	      \draw (acd) circle ({1 / 2 / sqrt(3)});
   	      \end{scope}
	      
	      \begin{scope}[canvas is plane={O(\pgfpointanchor{abd}{center})x(\pgfpointanchor{abdu}{center})y(\pgfpointanchor{abdv}{center})}]
	      \draw (abd) circle ({1 / 2 / sqrt(3)});
	      \draw[fill] (abd) circle ({1 / 40});
	      \ifthenelse{#2 = 0}
	      {\draw[very thick, cyan, ->] (barycentric cs:a=0.5,b=0.5) arc (-90:240:{1/2/sqrt(3)});}
	      {\draw[very thick, green, ->] (barycentric cs:a=0.5,b=0.5) arc (-90:-420:{1/2/sqrt(3)});}
	      ;
	      \end{scope}
	      
		\ifthenelse{#1 = 0}
		{
			\node (abdn) at (barycentric cs:abd=1,c=-0.37) {};
			\draw [thick,magenta,->,>=stealth] (abd) -- (abdn);
		}
		{
		}
		;
	}
	}
}

\tikzset{
	pics/tetra/.style n args={4}{
	code = { %
		\node (a) at ({1/sqrt(3)}, 0, 0) {};
		\node (b) at ({-1/2/sqrt(3)}, { 1/2}, 0) {};
		\node (c) at ({-1/2/sqrt(3)}, {-1/2}, 0) {};
		\node (d) at (0, 0, {1 * sqrt(2/3)}) {};
		
		\coordinate (abc) at (barycentric cs:a=1,b=1,c=1);
		\coordinate (abd) at (barycentric cs:a=1,b=1,d=1);
		\coordinate (bcd) at (barycentric cs:b=1,c=1,d=1);
		\coordinate (acd) at (barycentric cs:a=1,c=1,d=1);
	      
	      \coordinate (abcu) at ($(abc)+(a)-(b)$);
	      \coordinate (abcv) at (barycentric cs:abc={1-sqrt(3)},c={sqrt(3)});
	      \coordinate (abdu) at ($(abd)+(b)-(a)$);
	      \coordinate (abdv) at (barycentric cs:abd={1-sqrt(3)},d={sqrt(3)});
	      \coordinate (bcdu) at ($(bcd)+(c)-(b)$);
	      \coordinate (bcdv) at (barycentric cs:bcd={1-sqrt(3)},d={sqrt(3)});
	      \coordinate (acdu) at ($(acd)+(a)-(c)$);
	      \coordinate (acdv) at (barycentric cs:acd={1-sqrt(3)},d={sqrt(3)});
	      
	      \draw[] (a.center) -- (b.center);
	      \draw (b.center) -- (c.center);
	      \draw (c.center) -- (a.center);
	      \draw[] (a.center) -- (d.center);
	      \draw[] (b.center) -- (d.center);
	      \draw (c.center) -- (d.center);
	      
	      \begin{scope}[canvas is plane={O(\pgfpointanchor{abc}{center})x(\pgfpointanchor{abcu}{center})y(\pgfpointanchor{abcv}{center})}]
	      \draw (abc) circle ({1 / 2 / sqrt(3)});
	      \draw[fill] (abc) circle ({1 / 40});
	      \draw[very thick, cyan, ->] (barycentric cs:a=0.5,b=0.5) arc (-90:240:{1/2/sqrt(3)});
	      \end{scope}
	      	      
   	      \begin{scope}[canvas is plane={O(\pgfpointanchor{bcd}{center})x(\pgfpointanchor{bcdu}{center})y(\pgfpointanchor{bcdv}{center})}]
   	      \draw (bcd) circle ({1 / 2 / sqrt(3)});
	      \draw[fill] (bcd) circle ({1 / 40});
	      \draw[very thick, cyan, ->] (barycentric cs:b=0.5,c=0.5) arc (-90:240:{1/2/sqrt(3)});
   	      \end{scope}
   	      
   	      \begin{scope}[canvas is plane={O(\pgfpointanchor{acd}{center})x(\pgfpointanchor{acdu}{center})y(\pgfpointanchor{acdv}{center})}]
   	      \draw (acd) circle ({1 / 2 / sqrt(3)});
   	      \draw[fill] (acd) circle ({1 / 40});
  	      \draw[very thick, cyan, ->] (barycentric cs:a=0.5,c=0.5) arc (-90:240:{1/2/sqrt(3)});
   	      \end{scope}
	      
	      \begin{scope}[canvas is plane={O(\pgfpointanchor{abd}{center})x(\pgfpointanchor{abdu}{center})y(\pgfpointanchor{abdv}{center})}]
	      \draw (abd) circle ({1 / 2 / sqrt(3)});
	      \draw[fill] (abd) circle ({1 / 40});
	      \draw[very thick, cyan, ->] (barycentric cs:a=0.5,b=0.5) arc (-90:240:{1/2/sqrt(3)});
	      \end{scope}
	      
		\ifthenelse{#1 = 0}
		{
			\node (abcn) at (barycentric cs:abc=1,d=-0.37) {};
			\draw [thick,magenta,->,>=stealth] (abc) -- (abcn);
		}
		{
			\node (abcn) at (barycentric cs:abc=1,d=1.5) {};
			\draw [thick,orange,->,>=stealth] (abc) -- (abcn);
		}
		;

		\ifthenelse{#2 = 0}
		{
			\node (abdn) at (barycentric cs:abd=1,c=-0.37) {};
			\draw [thick,magenta,->,>=stealth] (abd) -- (abdn);
		}
		{
			\node (abdn) at (barycentric cs:abd=1,c=1.5) {};
			\draw [thick,orange,->,>=stealth] (abd) -- (abdn);
		}
		;
		
		\ifthenelse{#3 = 0}
		{
			\node (bcdn) at (barycentric cs:bcd=1,a=-0.37) {};
			\draw [thick,magenta,->,>=stealth] (bcd) -- (bcdn);
		}
		{
			\node (bcdn) at (barycentric cs:bcd=1,a=1.5) {};
			\draw [thick,orange,->,>=stealth] (bcd) -- (bcdn);
		}
		;

		\ifthenelse{#4 = 0}
		{
			\node (acdn) at (barycentric cs:acd=1,b=-0.37) {};
			\draw [thick,magenta,->,>=stealth] (acd) -- (acdn);
		}
		{
			\node (acdn) at (barycentric cs:acd=1,b=1.5) {};
			\draw [thick,orange,->,>=stealth] (acd) -- (acdn);
		}
		;
	}
	}
}

\tikzset{
	pics/fractal_generation/.style n args={2}{
	code = { %
	      \node (a) at ({1/sqrt(3)}, 0, 0) {};
	      \node (b) at ({-1/2/sqrt(3)}, { 1/2}, 0) {};
	      \node (c) at ({-1/2/sqrt(3)}, {-1/2}, 0) {};
	      \node (d) at (0, 0, {1 * sqrt(2/3)}) {};
	      
	      \coordinate (abc) at (barycentric cs:a=1,b=1,c=1);
	      \coordinate (abd) at (barycentric cs:a=1,b=1,d=1);
	      \coordinate (bcd) at (barycentric cs:b=1,c=1,d=1);
	      \coordinate (acd) at (barycentric cs:a=1,c=1,d=1);
	      
		\ifthenelse{#1 = 0}
		{
		}
		{
			\node (abdn) at (barycentric cs:abd=1,c=1.5) {};
			\draw [thick,orange,->,>=stealth] (abd) -- (abdn);
		}
		;
	      
	      \coordinate (abcu) at ($(abc)+(b)-(a)$);
	      \coordinate (abcv) at (barycentric cs:abc={1-sqrt(3)},c={sqrt(3)});
	      \coordinate (abdu) at ($(abd)+(b)-(a)$);
	      \coordinate (abdv) at (barycentric cs:abd={1-sqrt(3)},d={sqrt(3)});
	      \coordinate (bcdu) at ($(bcd)+(c)-(b)$);
	      \coordinate (bcdv) at (barycentric cs:bcd={1-sqrt(3)},d={sqrt(3)});
	      \coordinate (acdu) at ($(acd)+(c)-(a)$);
	      \coordinate (acdv) at (barycentric cs:acd={1-sqrt(3)},d={sqrt(3)});
	      
	      \draw[thick] (a.center) -- (b.center);
	      \draw (b.center) -- (c.center);
	      \draw (c.center) -- (a.center);
	      \draw[thick] (a.center) -- (d.center);
	      \draw[thick] (b.center) -- (d.center);
	      \draw (c.center) -- (d.center);
	      
	      \begin{scope}[canvas is plane={O(\pgfpointanchor{abc}{center})x(\pgfpointanchor{abcu}{center})y(\pgfpointanchor{abcv}{center})}]
	      \draw (abc) circle ({1 / 2 / sqrt(3)});
	      \end{scope}
	      	      
   	      \begin{scope}[canvas is plane={O(\pgfpointanchor{bcd}{center})x(\pgfpointanchor{bcdu}{center})y(\pgfpointanchor{bcdv}{center})}]
   	      \draw (bcd) circle ({1 / 2 / sqrt(3)});
   	      \end{scope}
   	      
   	      \begin{scope}[canvas is plane={O(\pgfpointanchor{acd}{center})x(\pgfpointanchor{acdu}{center})y(\pgfpointanchor{acdv}{center})}]
   	      \draw (acd) circle ({1 / 2 / sqrt(3)});
   	      \end{scope}
	      
	      \begin{scope}[canvas is plane={O(\pgfpointanchor{abd}{center})x(\pgfpointanchor{abdu}{center})y(\pgfpointanchor{abdv}{center})}]
	      \draw (abd) circle ({1 / 2 / sqrt(3)});
	      \draw[fill] (abd) circle ({1 / 40});
	      \ifthenelse{#2 = 0}
	      {\draw[very thick, cyan, ->] (barycentric cs:a=0.5,b=0.5) arc (-90:240:{1/2/sqrt(3)});}
	      {\draw[very thick, green, ->] (barycentric cs:a=0.5,b=0.5) arc (-90:-420:{1/2/sqrt(3)});}
	      ;
	      \end{scope}
	      
		\ifthenelse{#1 = 0}
		{
			\node (abdn) at (barycentric cs:abd=1,c=-0.37) {};
			\draw [thick,magenta,->,>=stealth] (abd) -- (abdn);
		}
		{
		}
		;
	}
	}
}

\title{Modeling and Experimental Validation of a Fractal Tetrahedron UAS Assembly}
\author{
K\'evin Garanger \\
School of Aerospace Engineering\\
Georgia Institute of Technology\\
Atlanta, Georgia 30332, USA\\
kevin.garanger@gatech.edu
\and
Jeremy Epps \\
School of Aerospace Engineering\\
Georgia Institute of Technology\\
Atlanta, Georgia 30332, USA\\
jeremy.epps@gatech.edu
\and
Eric Feron\\
Computer, Electrical and Mathematical Sciences \& Engineering\\
King Abdullah University of Science and Technology\\
Thuwal 23955, Saudi Arabia\\
eric.feron@kaust.edu.sa
}
\date{}

\begin{document}

\maketitle

\begin{abstract}
   This paper presents the foundation of a modular robotic system comprised of several novel modules in the shape of a tetrahedron. Four single-propeller submodules are assembled to create the Tetracopter, a tetrahedron-shaped quad-rotorcraft used as the elementary module of a modular flying system. This modular flying system is built by assembling the different elementary modules in a fractal shape. The fractal tetrahedron structure of the modular flying assembly grants the vehicle more rigidity than a conventional two-dimensional modular robotic flight system while maintaining the relative efficiency of a two-dimensional modular robotic flight system. A prototype of the Tetracopter has been modeled, fabricated, and successfully flight-tested by the Decision and Control Laboratory at the Georgia Institute of Technology. The results of this research set the foundation for the development of Tetrahedron rotorcraft that can maintain controllable flight and assemble in flight to create a Fractal Tetrahedron Assembly.
\end{abstract}

\section{Introduction}
Over the past two decades, a consequence of the technological advances in lithium battery and the miniaturization of inertial measurement units has been the emergence of small multi-rotor aircraft. The majority of these aircraft are structured so that their rotors lie on the same plane. This is done for efficiency in flight, control, and simplicity in design and fabrication. Recent works have studied the possibility of combining several multi-rotor aircraft in a rigid flying structure~\cite{8461014, 5509882, 8741685}. The main purpose of such a structure would be to use the combined lifting power of all of its members to transport heavier payloads. Other advantages of modular flying structure is the ability to put in common the capabilities of its members which could have, for instance, different sensing capabilities. One can also imagine a system in which a transfer of energy would occur between members with high battery reserves and members with almost depleted batteries. Creating a flying assembly using an array of rotorcraft becomes a disadvantage in the face of external forces that could cause deformations and a loss of efficiency as the assembly grows in size or creates high internal stress leading to structural mechanical failure. Examples of high loading that can occur on the structure of a flight array are during the lifting of a heavy payload or when flying in turbulent air. This paper presents the preliminary design to a novel multi-rotor modular robotic flight system whose mechanical properties and efficiency do not suffer from the increase in the number of modules.  

\begin{figure}[h]
    \centering
    \begin{subfigure}[t]{0.22\textwidth}
        \includegraphics[width=\textwidth]{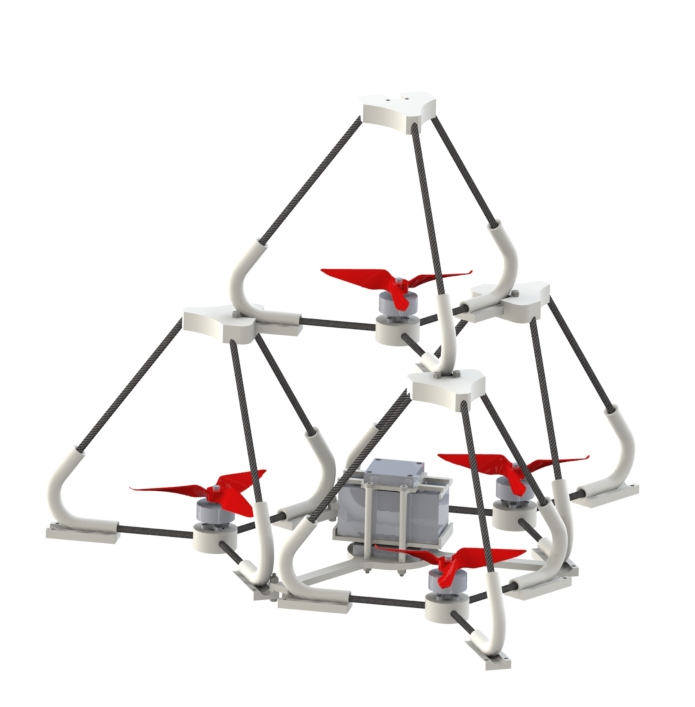}
        \caption{Model of the Tetracopter.}
        \label{fig:elem_mod_2d}
    \end{subfigure}
    \begin{subfigure}[t]{0.25\textwidth}
        \includegraphics[width=\textwidth]{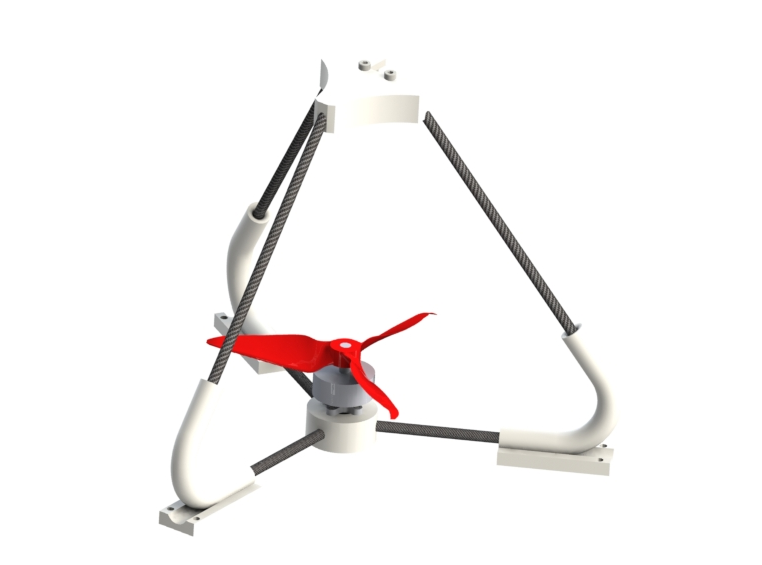}
        \caption{Model of a Tetracopter single-propeller submodule.}
        \label{fig:submodule}
    \end{subfigure}
    \caption{Models used for the prototyping of the Tetracopter elementary module.}
\end{figure}

The adverse effects of the increasing size of the assembly are mitigated by using a fractal-inspired spatial configuration of the multiple rotorcraft. A single rotorcraft is called an elementary module and several elementary modules can assemble at different scales to form a Fractal Tetrahedron Assembly (FTA) based on the Sierpinski tetrahedron~\cite{frame2016fractal}.
The goal of the FTA design is to provide rigidity in three-dimensions while maintaining the efficiency of a virtually two-dimensional modular robotic flight system. Different designs of the elementary module can be considered. In this work, we consider an elementary module consisting of a quad-rotorcraft with a regular tetrahedron structure, as seen in Fig.~\ref{fig:elem_mod_2d}. In practice, this tetrahedral quad-rotorcraft, which we name Tetracopter, is made of four identical submodules and a common payload including the battery and electronics of the quad-rotorcraft. Fig.~\ref{fig:submodule} shows this submodule, which has only one rotor and is therefore uncontrollable by itself. The results of this research are the foundation of future research of Georgia Tech's Decision and Control Laboratory that aims to develop tetrahedral rotorcraft that can self-assemble in flight.

This paper is organized as follows:
\begin{itemize}
    \item The details of the Fractal Tetrahedron Assembly are presented in Section~\ref{sec:assembly}, along with the analyses and results that justify this assembly scheme. Among them are a study of the dynamics of the assembly and some simulated results of the internal forces of the assembly structure in different situations.
    \item The Tetracopter, which is used as the elementary module of the FTA in most of the results presented here, is modeled in Section~\ref{sec:tetra_model}. The dynamics of the Tetracopter are derived and the effects of the vehicle's frame on the inflow of the rotors are discussed.
    \item The prototype of the Tetracopter is presented in Section~\ref{sec:prototype}.
    \item In Section~\ref{sec:fault_tol}, the tolerance of the FTA to motor failures is assessed.
    \item Directions for future work that are suggested or will be explored by the authors are presented in Section~\ref{sec:future_work}.
    \item Finally, a summary of the findinds of this work is given in the conclusion found in Section~\ref{sec:conclusion}.
\end{itemize}

\subsection{Related Works}
The field of modular robotic systems has been of interest to the robotics community for over a decade but is still a rich research field. This field has a number of interesting problems ranging from the hardware used to assemble modular robotic systems, the necessary control methods to gather the robots in a synchronized manner, as well as the change in control law necessary for the individual robots to work as one united system once assembled. Ahmadzadeh and Masehian discuss these  problems and their potential solutions in detail \cite{ahmadzadeh_masehian_2015}. There has also been previous research that delves into the connection between fractals and the assembly of modular robots \cite{bie}. Research conducted at ETH Zurich has shown that individual robots that are not controllable can be assembled into a two-dimensional Distributed Flight Array to form a system that sustains controllable flight \cite{5509882}. An ongoing project by NASA's JPL studies the concept of flying amphibious robots capable of assembling to form different objects \cite{shapeshifter}. Although these works have similar elements as the presented research, they do not involve flight vehicles that are assembled in three dimensions nor do they address that modular robotic systems provide structural rigidity as the Fractal Tetrahedral Assembly does.

\section{Self-assembly of multiple modules}\label{sec:assembly}

\subsection{The Sierpinski tetrahedron architecture}
The assembly pattern of the Fractal Tetrahedron UAS Assembly is based on the Sierpinski tetrahedron \cite{frame2016fractal}, a fractal with an overall shape similar to a regular tetrahedron.
This structure is chosen for several reasons:
\begin{itemize}
    \item The possibility to assemble several building blocks in a 3-dimensional fashion as opposed to a flat layout, thus improving the structural properties of the assembly;
    \item The main building block, a tetrahedron, is suitable for the design of an autonomous flying module as shown in section \ref{sec:tetra_model};
    \item The fractal structure in a tetrahedron shape creates relatively small structural loads in the bottom modules. Moreover, this loading scales well as combining four assemblies only multiplies the loading by a factor of $4/3$;
    \item When the 2D-Tetracopter is used as the elementary module, the rotors never overlap with one another.
\end{itemize}

\subsection{Generative rule of the fractal assembly}
The generation of the fractal assembly is done recursively from an elementary module using the generation rule described below.

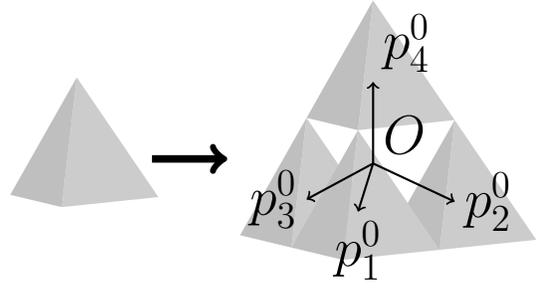
\begin{figure}[h]
	\centering
	\begin{tikzpicture}
		\draw[->, line width=1mm] (1, 0) -- (2, 0);
		\tdplotsetmaincoords{85}{100}
		\begin{scope}[tdplot_main_coords, scale=2]
			\sierpinskigasket[0]{0};
		\end{scope};
	\begin{scope}[tdplot_main_coords, scale=4, shift=({0, 1, 0}), font=\huge]
		\sierpinskigasket[1]{1};
	\end{scope}
	\end{tikzpicture}
	\caption{Generation of the 1-assembly from the elementary module.}
	\label{fig:fractal_generation_01}
\end{figure}

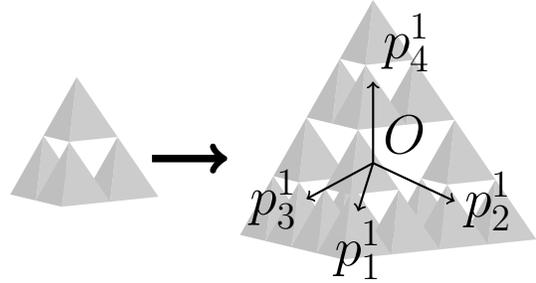
\begin{figure}[h]
	\centering
	\begin{tikzpicture}
		\draw[->, line width=1mm] (1, 0) -- (2, 0);
		\tdplotsetmaincoords{85}{100}
		\begin{scope}[tdplot_main_coords, scale=2]
			\sierpinskigasket[0]{1};
		\end{scope};
	\begin{scope}[tdplot_main_coords, scale=4, shift=({0, 1, 0}), font=\huge]
		\sierpinskigasket[1]{2};
	\end{scope}
	\end{tikzpicture}
	\caption{Generation of the 2-assembly from the 1-assembly.}
	\label{fig:fractal_generation_12}
\end{figure}

Let $\mathcal{T}$ be a regular tetrahedron of edge length $l$. The distance from the center of $\mathcal{T}$ to its vertices is defined as $r = \frac12\sqrt{\frac32}l$.
For $i \in \{1,\dots, 4\}$, let $\bm{p}_i $ be the unit vector pointing from the center $O$ of $\mathcal{T}$ to the vertex $v_i$ as shown in Fig.~\ref{fig:fractal_generation_01}.

Let $\mathcal{B}$ be a rigid body used as the elementary module of the fractal assembly. The 0-assembly is defined to be equal to the elementary module. For $n\geq 0$, The (n+1)-assembly is defined as the rigid body made from four identical n-assemblies $\mathcal{A}_n^1$, $\mathcal{A}_n^2$, $\mathcal{A}_n^3$, and $\mathcal{A}_n^4$ translated respectively by $\bm{p}_1^{n+1}$, $\bm{p}_2^{n+1}$, $\bm{p}_3^{n+1}$, and $\bm{p}_4^{n+1}$ and connected together with rigid joints. $\bm{p}_i^{n+1}$ is defined by $\bm{p}_i^{n+1} = 2^n r \bm{p}_i$.

With this definition, the elementary module can be of any shape. However, in practice, the shape is constrained so that one can find enough connection points to connect all the elementary modules without having them overlapping. Another desirable property of the assembly that can be sought is that it respects the rotational invariance of the tetrahedron. As shown in~\cite{kishimoto2006hierarchical}, it is possible to create such an assembly from the five regular convex polyhedra. It would be therefore possible to adapt this work with a different elementary module, such as a flying cube or octahedron.

\subsection{Propeller disk area of the Tetracopter fractal assembly}

\begin{figure}[h!]
    \centering
    \begin{subfigure}[b]{0.15\textwidth}
        \includegraphics[width=\textwidth]{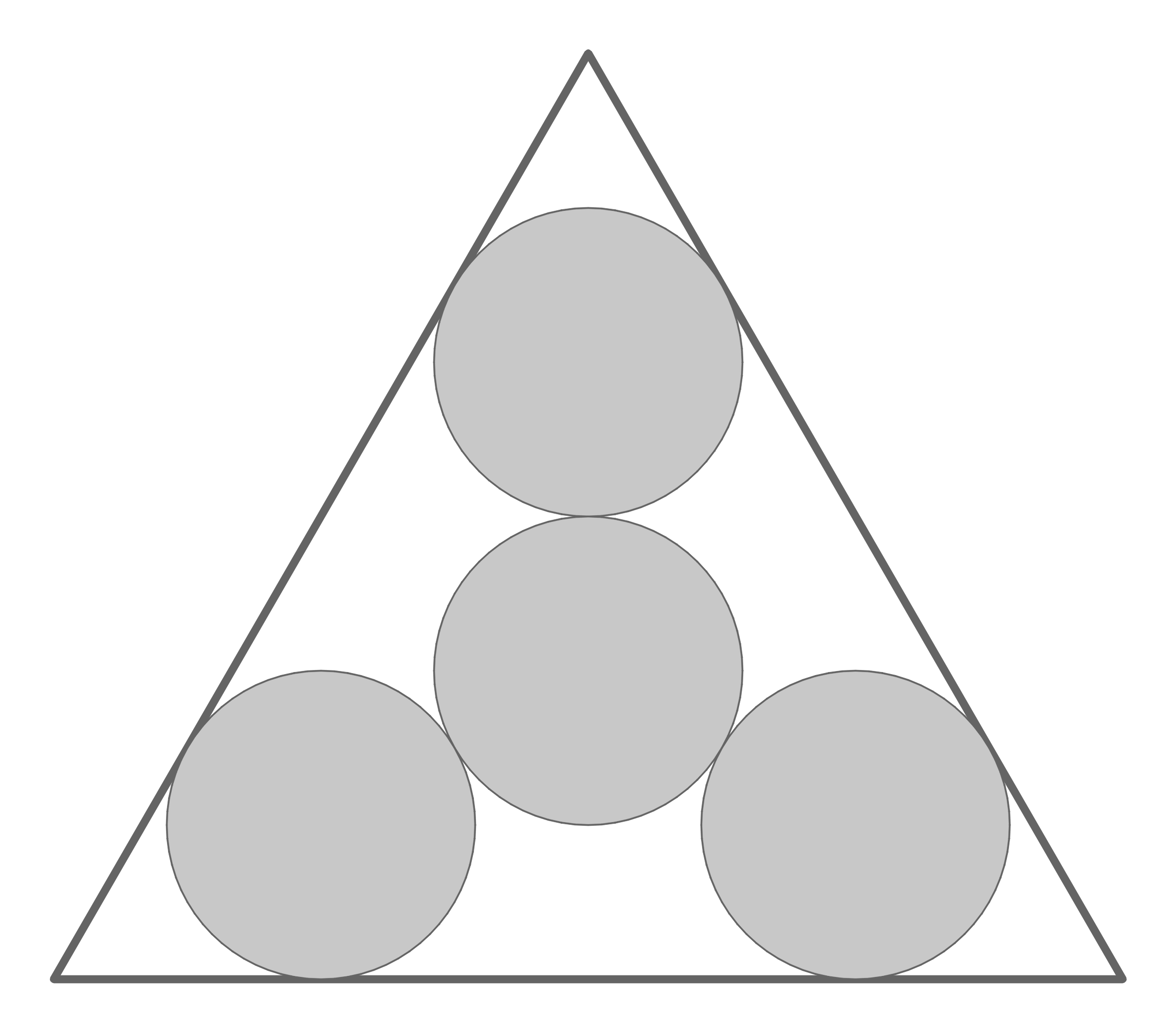}
        \caption{$4^1 = 4$ rotors}
    \end{subfigure}
    \begin{subfigure}[b]{0.15\textwidth}
        \includegraphics[width=\textwidth]{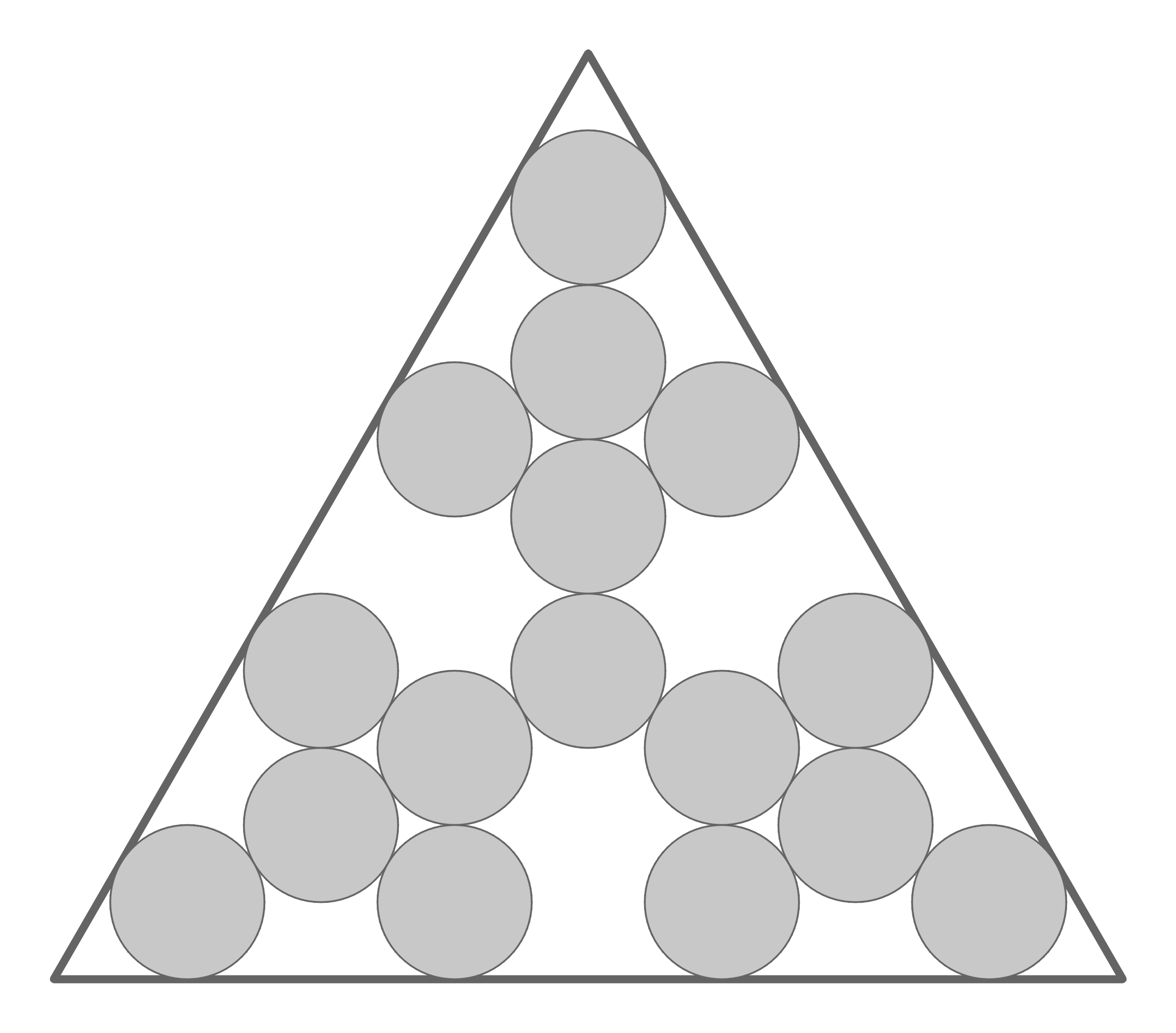}
        \caption{$4^2 = 16$ rotors}
    \end{subfigure}
    \begin{subfigure}[b]{0.15\textwidth}
        \includegraphics[width=\textwidth]{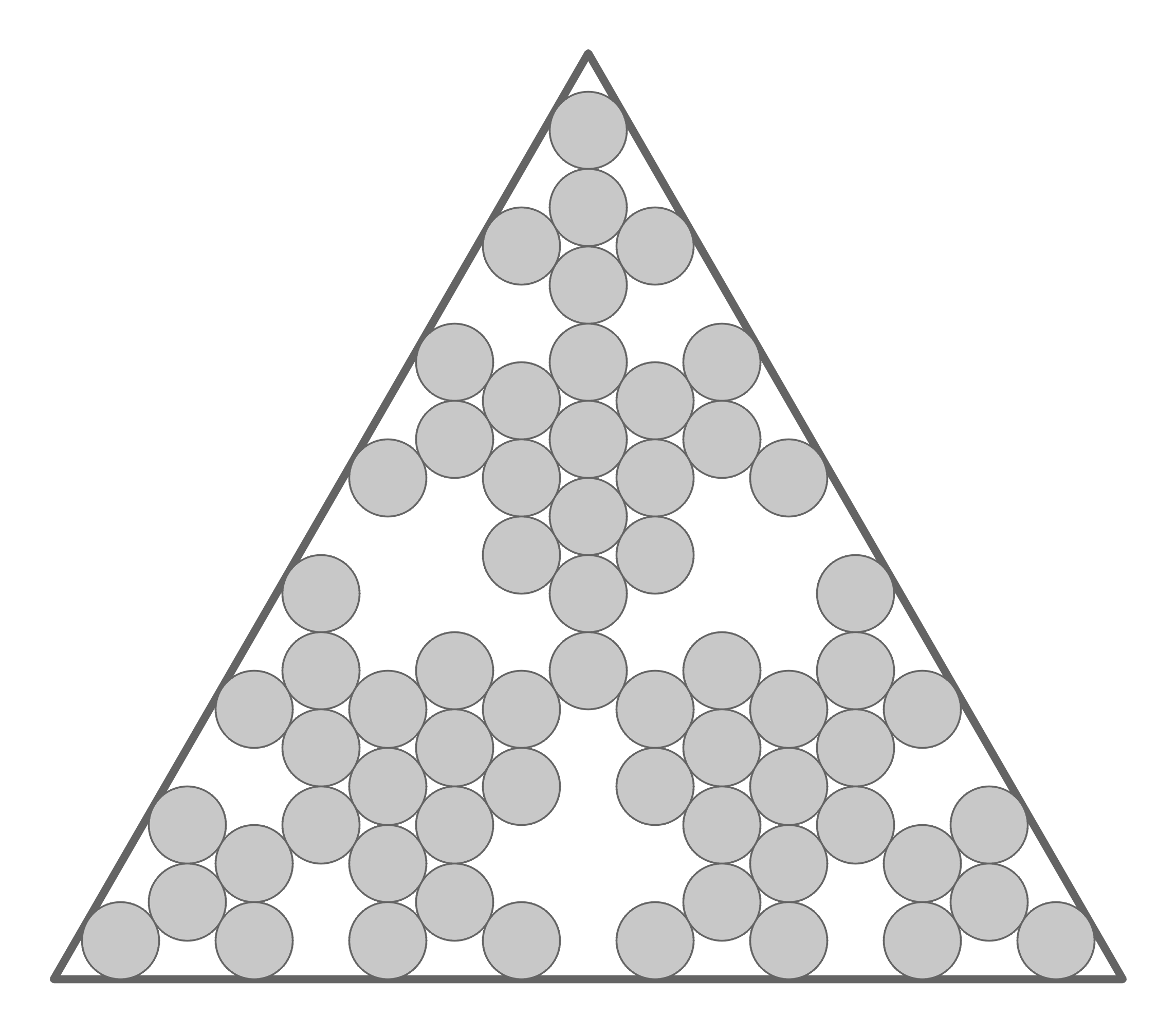}
        \caption{$4^3 = 64$ rotors}
    \end{subfigure}
    \begin{subfigure}[b]{0.15\textwidth}
        \includegraphics[width=\textwidth]{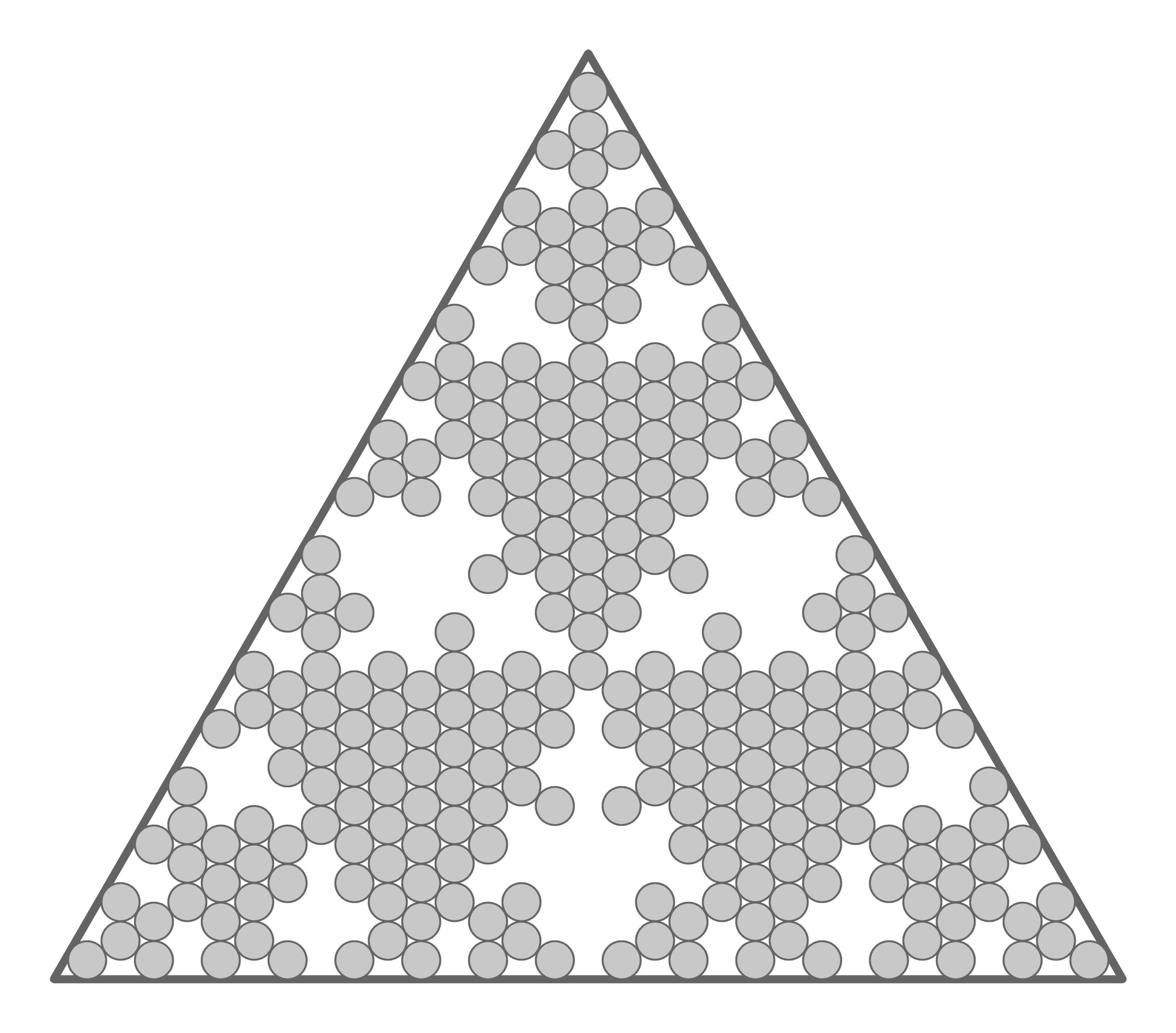}
        \caption{$4^4 = 256$ rotors}
    \end{subfigure}
    \begin{subfigure}[b]{0.15\textwidth}
        \includegraphics[width=\textwidth]{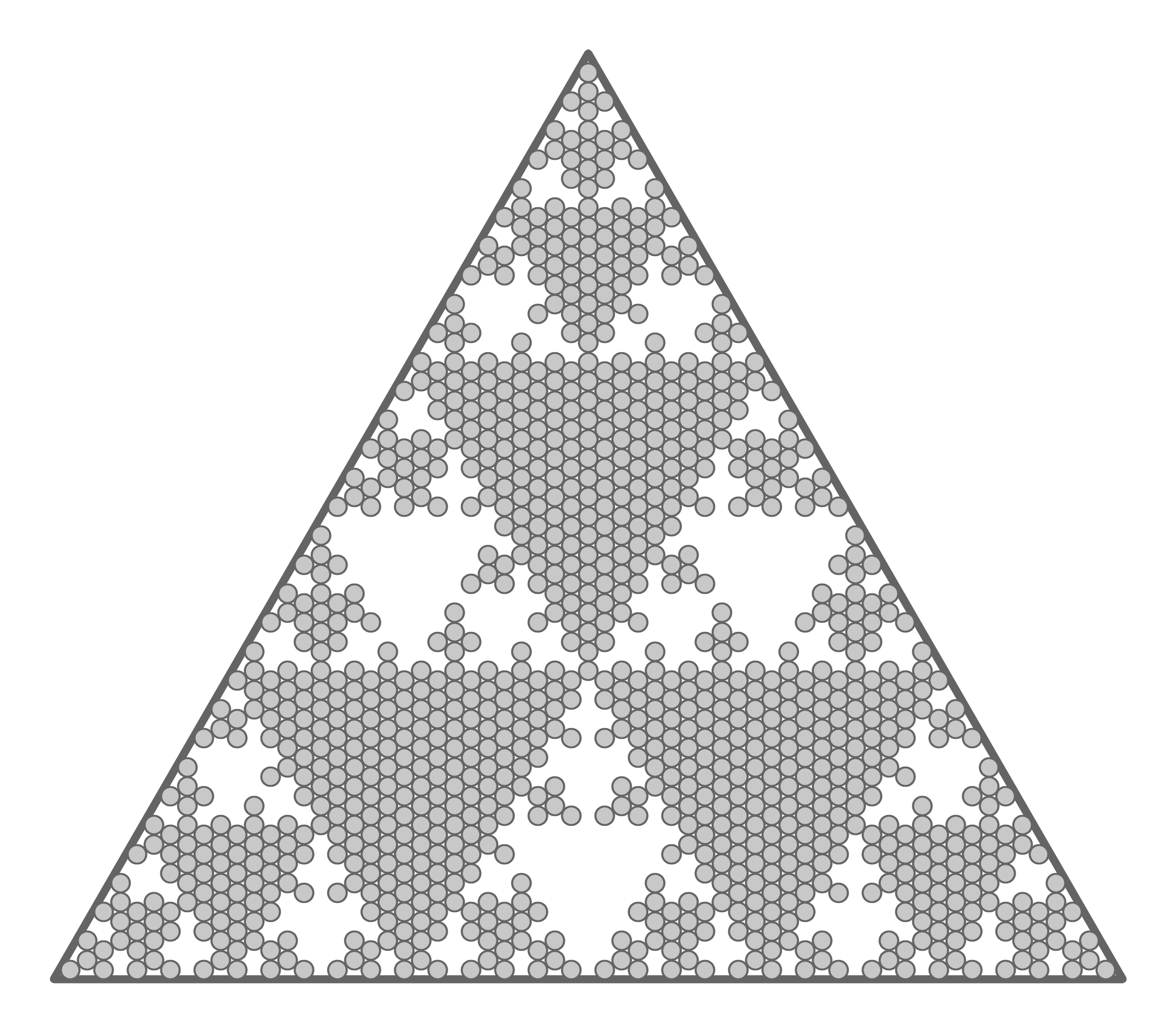}
        \caption{$4^5 = 1024$ rotors}
    \end{subfigure}
    \begin{subfigure}[b]{0.15\textwidth}
        \includegraphics[width=\textwidth]{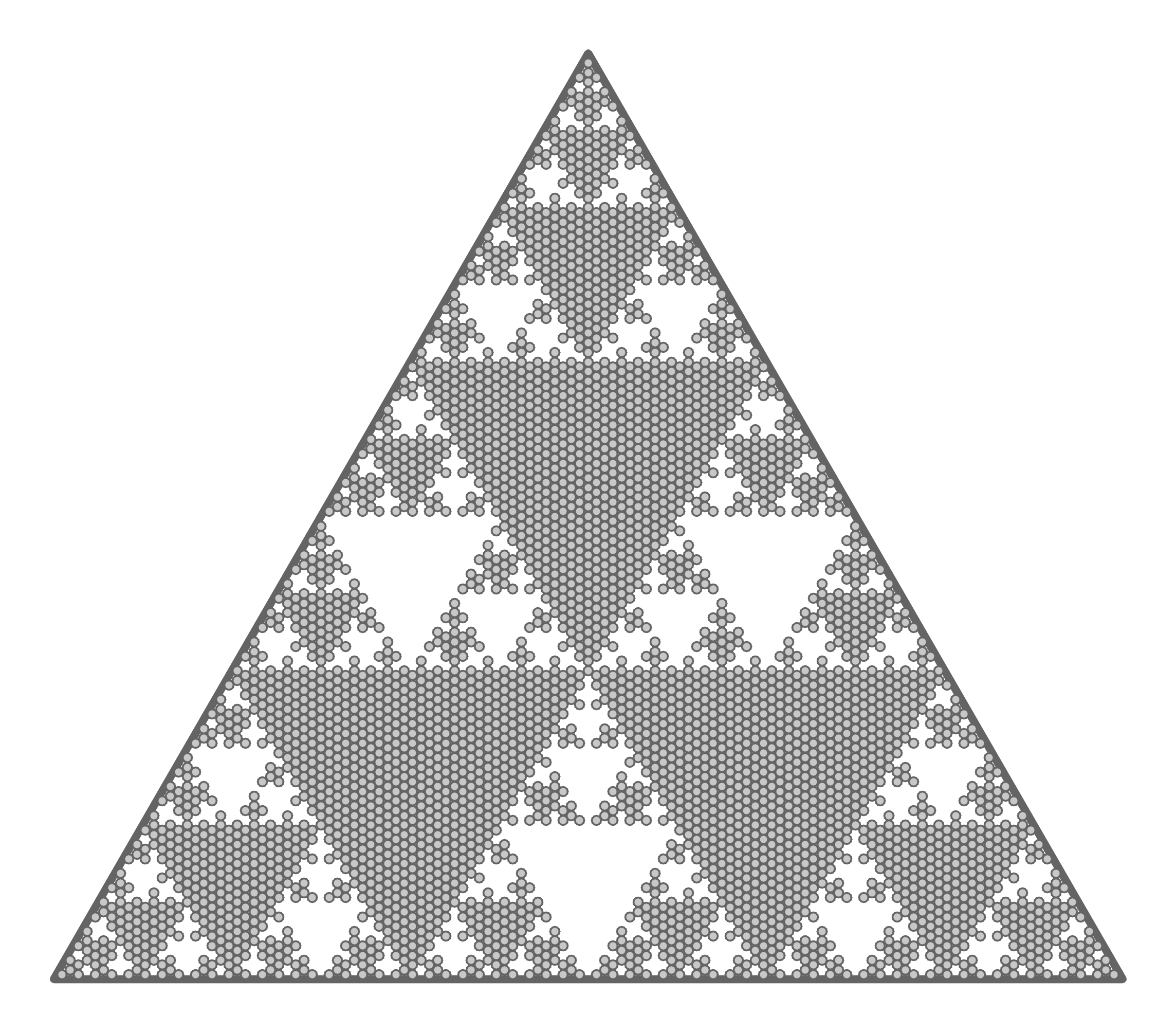}
        \caption{$4^6 = 4096$ rotors}
    \end{subfigure}
    \caption{Projected view of the area covered by the rotors of the fractal tetrahedron assembly at different levels.}
    \label{fig:rotors_top_view}
\end{figure}

An important consideration of the fractal tetrahedron design used in conjunction with the Tetracopter elementary module is that no rotor is ever directly above any other rotor. It can indeed be shown that the projection of the area covered by all the rotors on the horizontal plane is a set of disks that do not overlap as shown on figure~\ref{fig:rotors_top_view}.

The total area $\mathcal{A_R}$ covered by the rotors is therefore the sum of all the disks areas. The ratio of $\mathcal{A_R}$ by $\mathcal{A_B}$, with $\mathcal{A_B}$ being the area of the base of the assembly, is the same at every scale of the assembly. Indeed, combining $4$ assemblies multiplies both the rotor area and the base area by $4$. Therefore $\mathcal{A_R}$ depends only on the size of the fractal tetrahedron assembly but not on the size of its elementary modules. If we define $a$ as the length of a side of the triangular base, the rotor area is equal to the area of the inscribed circle of the base:
\begin{equation}
    \begin{split}
        \mathcal{A_R} = \frac{\pi}{12}a^2.
    \end{split}
\end{equation}
On the other hand, the area of the triangular base is given by 
\begin{equation}
    \mathcal{A_B} = \frac{\sqrt{3}}{4}a^2.
\end{equation}
The ratio of the total disc area by the area covered by the base of the assembly is therefore equal to
\[
    \frac{\pi}{3\sqrt{3}}.
\]
In comparison, the maximum disk occupancy achievable on the plane is equal to
\[
    \frac{2\pi}{3\sqrt{3}}.
\]
This value would correspond to tiling the plane with regular hexagons and considering the inscribed circle of each hexagon.

\subsection{Inertia of the fractal assembly}

Let $m_n$ and $J_n$ be respectively the mass and the inertia tensor of the n-assembly for $n \geq 0$. The subscript is omitted for $n=0$, such that $m$ and $J$ are the mass and inertia tensor of the elementary module. By applying the parallel axis theorem on each inertia tensor of the $4$ components of the (n+1)-assembly and adding them, we obtain the following formula:
\begin{align*}
J_{n+1} &= \sum_{i=1}^{4} J_n + m_n \left((\bm{p}_i^{n+1} \cdot \bm{p}_i^{n+1}) I_3 - {\bm{p}_i^{n+1}}^T \otimes \bm{p}_i^{n+1} \right) \\
&= 4J_n + 4^n m_n r^2 \left(4I_3 - M\right).
\end{align*}
where $\otimes$ is the Kronecker product and
\begin{equation*}
    M = \sum_{i=1}^{4} \bm{p}_i \otimes \bm{p}_i = \frac{4}{3} I_3.
\end{equation*}

Since $m_n = 4^n m$,
we can write
\begin{align*}
    J_{n+1}&\ = 4J_n + \frac{8}{3}16^n m r^2 I_3 \\
\frac{J_{n+1}}{16^{n+1}}&\ = \frac{1}{4}\frac{J_n}{16^n} + \frac{1}{6}mr^2_3 \\
\frac{J_{n+1}}{16^{n+1}}&\ - \frac{2}{9} m r^2 I_3 \\
&\ =\frac{1}{4}\left(\frac{J_n}{16^n} - \frac{2}{9}mr^2I_3\right) \\
&\ = \frac{1}{4^{n+1}}\left(J - \frac{2}{9}mr^2I_3\right)
\end{align*}
which finally gives
\begin{align}\label{eq:fractal_inertia}
J_{n+1} =\ &\frac{2}{9}16^{n+1}m r^2 I_3 \nonumber \\
&+ 4^{n+1}\left(J - \frac{2}{9}m r^2 I_3\right).
\end{align}

\subsection{Dynamics of the fractal assembly}

The linearized dynamics of the fractal assembly are derived recursively. The derivation is made for an arbitrary elementary module with four propellers under the assumption that the center of mass of each assembly is at the center of the tetrahedron it forms.

Let us consider four identical n-assemblies each made of $4^n$ elementary modules and $4^{n+1}$ propellers that are combined to make a (n+1)-assembly. For each n-assembly considered separately and indexed by $i \in \{1\dots 4\}$, we can consider the forces and moments acting on its center of mass in the body-fixed frame for a given vector of controls $\bm{u}^i$ of size $4^{n+1}$.
We write $\bm{T}_i$ the total thrust it produces, $\bm{M}_i^T$ the torques induced on its center of mass by differential thrust, and $\bm{M}_i^C$ the torque induced by the rotation of the assembly rotors (gyroscopic effects and counteracting torques). The equivalent quantities for the (n+1)-assembly are given by $\bm{T}$, $\bm{M}^T$, and $\bm{M}^C$, respectively. The concatenation of the four control vector $u_i$ is the control vector of the (n+1)-assembly, given by
\[
    \bm{u} = 
    \begin{bmatrix}
        \bm{u}^1 \\ \bm{u}^2 \\ \bm{u}^3 \\ \bm{u}^4
    \end{bmatrix}.
\]
The attitude of the four n-assemblies and of the (n+1)-assembly that they constitute is shared since the structure is assumed rigid. This attitude is given by the vector of angular velocities written $\bm{\Omega}$.
Those terms are the only ones necessary for the derivation of the linearized dynamics of the fractal assembly since all the other terms vanish during the linearization. When combining the four n-assemblies, the following equalities hold:
\begin{equation}\label{eq:fractal_dyn:recursion}
    \begin{split}
        \bm{T} &= \sum_{i=1}^{4} \bm{T}_i \\
        \bm{M}_T &= \sum_{i=1}^{4} \bm{p}_i^{n+1} \times \bm{T}_i + \bm{M}_i^T \\
        \bm{M}_C &= \sum_{i=1}^{4} \bm{M}_i^C.
    \end{split}
\end{equation}

Moreover, for each n-assembly $i$ and for a small change in $\bm{\Omega}$ and $\bm{u}^i$, the change in thrust and moment can be quantified by introducing the matrices $M_n^a(\bm{u}^i)$, $M_n^b(\bm{u}^i)$, $M_n^c(\bm{u}^i)$, and $M_n^d(\bm{u}^i)$, which come from the linearization of the forces and moments described previously around the state $\bm{\Omega}$ and controls $\bm{u}^{i}$. This linearization can be written
\begin{equation}\label{eq:fractal_dyn:lin}
    \begin{split}
        \Delta \bm{T}_i &=  M_n^a(\bm{u}^{i}) \Delta \bm{u}^i \\
        \Delta \bm{M}_i^T &= M_n^b(\bm{u}^{i}) \Delta \bm{u}^i \\
        \Delta \bm{M}_i^C &= M_n^c(\bm{u}^{i}, \bm{\Omega}) \Delta \bm{u}^i + M_n^d(\bm{u}^i, \bm{\Omega}) \Delta \Omega.
    \end{split}
\end{equation}

Combining \eqref{eq:fractal_dyn:recursion} and \eqref{eq:fractal_dyn:lin} provides recursive relations between the matrices $M_{n+1}^a$, $M_{n+1}^b$, $M_{n+1}^c$, $M_{n+1}^d$ and $M_n^a$, $M_n^b$, $M_n^c$, $M_n^d$ around a chosen equilibrium point. The dependency of theses matrices to the equilibrium attitude and controls are omitted for brevity.
For the specific case where $\bm{u}_i = u_0\mathbf{1}_{4^{n+1}}$ for every $i \in \{1\dots 4\}$, these recursive relations can be written
\begin{equation}
    \begin{split}
        M_{n+1}^a =&\ \mathbf{1}_4 \otimes M_n^a \\
        M_{n+1}^b =&\
        2^n r 
        \left[
            [\bm{p}_1]_\times | [\bm{p}_2]_\times | [\bm{p}_3]_\times | [\bm{p}_4]_\times
        \right]\left(I_4 \otimes M_n^a\right) \\ 
                           &\ +	
                           \mathbf{1}_4 \otimes M_n^b \\
        M_{n+1}^c =&\ \mathbf{1}_4 \otimes M_n^c \\
        M_{n+1}^d =&\ \mathbf{1}_4 \otimes M_n^d
    \end{split}
\end{equation}
where $\mathbf{1}_p = \left[1 1 \dots 1\right]$ is a matrix of size $1\times p$ for $p \geq 0$ and $[\bm{p}]_\times$ is the matrix corresponding to the left cross-product by $\bm{p}$.

Finally, by defining
\begin{equation}\label{eq:fractal_dyn:Q}
    Q = \left[
        [p_1]_\times M_0^a | [p_2]_\times M_0^a | [p_3]_\times M_0^a | [p_4]_\times M_0^a
    \right]
\end{equation}
we can write
\begin{equation}
    \begin{split}
        M_{n}^a =&\ \mathbf{1}_{4^n} \otimes M_0^a \\
        M_{n}^b =&\ \mathbf{1}_{4^n} \otimes M_0^b \\
                 &\ + r Q \sum_{k=0}^{n-1} 2^k \left( \mathbf{1}_{4^{n-k}} \otimes I_4 \otimes \mathbf{1}_{4^k} \otimes I_4 \right) \\
        M_{n}^c =&\ \mathbf{1}_{4^n}  \otimes M_0^c \\
        M_{n}^d =&\ \mathbf{1}_{4^n}  \otimes M_0^d.
    \end{split}
\end{equation}

The matrices $M_0^a$, $M_0^b$, $M_0^c$, $M_0^d$ are given by a linearization of the dynamics of the elementary module. The derivation of the dyanmics is made in the case of the Tetracopter in section~\ref{sec:tetra_model}. 
One can notice that the norms of $M_n^a$, $M_n^c$, and $M_n^d$ are proportional to $4^n$ while the dominant term in $M_n^b$ grows proportionally to $8^n$. Therefore, by looking at the moments equation for the n-assembly in which quadratic terms are omitted:
\begin{align*}
    J_n\dot{\bm{\Omega}} &\approx \bm{M}_T + \bm{M}_C,
\end{align*}
if the inertia is replaced using \eqref{eq:fractal_inertia}, and if the moments are replaced by their linearization around an equilibrium point with only the dominant terms as $n$ grows kept,
\begin{equation}
    \begin{split}
    &\frac{2}{9}16^{n}m r^2\left(4I_3 - M\right) \dot{\bm{\Omega}} \\ &\ \approx r Q \sum_{k=0}^n 2^k \left( \mathbf{1}_{4^{n-k}} \otimes I_4 \otimes \mathbf{1}_{4^k} \otimes I_4 \right) \Delta u
    \end{split}
\end{equation}
and one can see the left side of the equation grows proportionally to $16^n$ while the right side grows with $8^n$. As $n$ grows bigger, the dynamical system linearized equations can be seen as an expanded version of the reduced dynamical system and the system is expected to roughly behave similarly, with a characteristic time constant multiplied  by $2^n$.

\subsection{Internal forces of the fractal assembly}

The analysis of the internal forces of the fractal assembly structure can be made by representing the fractal assembly of $4^n$ tetrahedra as  a truss of $m_n$ members and $j_n$ joints. Since each tetrahedron is made of $6$ members, $m_n = 6\times4^n$ while the number of joints, found by induction on the size of the assembly, is given by $j_n = 2(4^n+1)$.

Therefore we have
\[
    m_n + 6 - 3j_n = 0
\]
and the truss is statically determinate internally. The direct stiffness method can be applied to evaluate the internal loads applied on each member in different scenarios. A similar analysis is done on the two dimensional Sierpinski gasket truss in \cite{kishimoto2000basic}. The authors of the mentioned publication show that the members internal forces of the truss do not differ much from the forces present in the complete triangle tessellation truss of same dimensions.

In the three dimensional cases, different simulations are performed to evaluate the rigidity of the assembly and the potential structural failures that could occur. The truss members parameters come from the type of tube used for the construction of the Tetracopter frame. 
The loads on the 2-assembly are computed with the direct stiffness method in three different scenarios:
\begin{itemize}
    \item At rest lying on the plane only supporting its own weight;
    \item Hovering a payload attached to the top elementary module;
    \item Hovering a payload attached to three different attachment points at each of the bottom corners of the assembly.
\end{itemize}

\begin{figure}[h!]
    \centering
    \begin{subfigure}[b]{0.45\textwidth}
        \includegraphics[width=\textwidth]{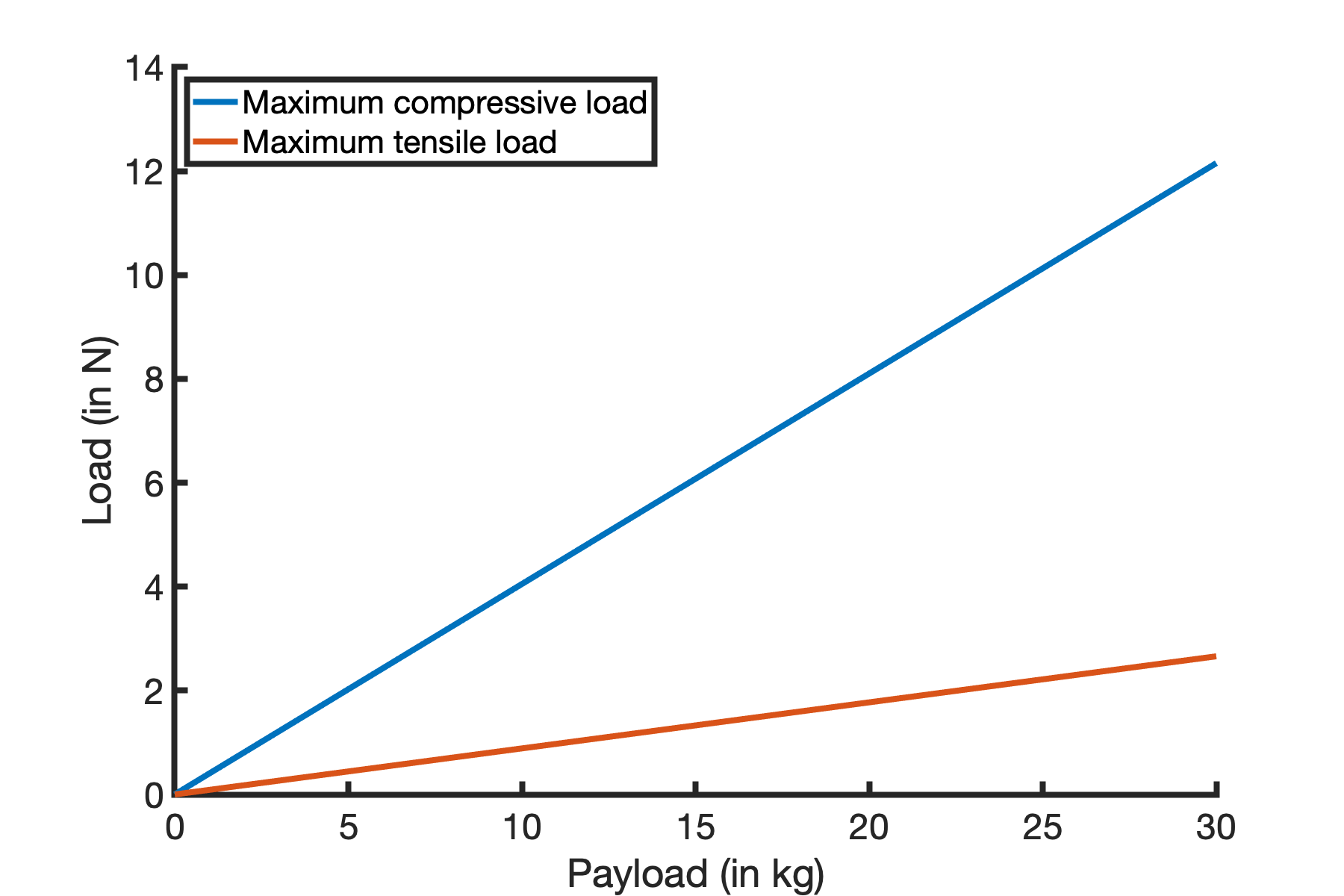}
        \caption{Single point top attachment scenario}
        \label{fig:loads:single_pt}
    \end{subfigure}
    \begin{subfigure}[b]{0.45\textwidth}
        \includegraphics[width=\textwidth]{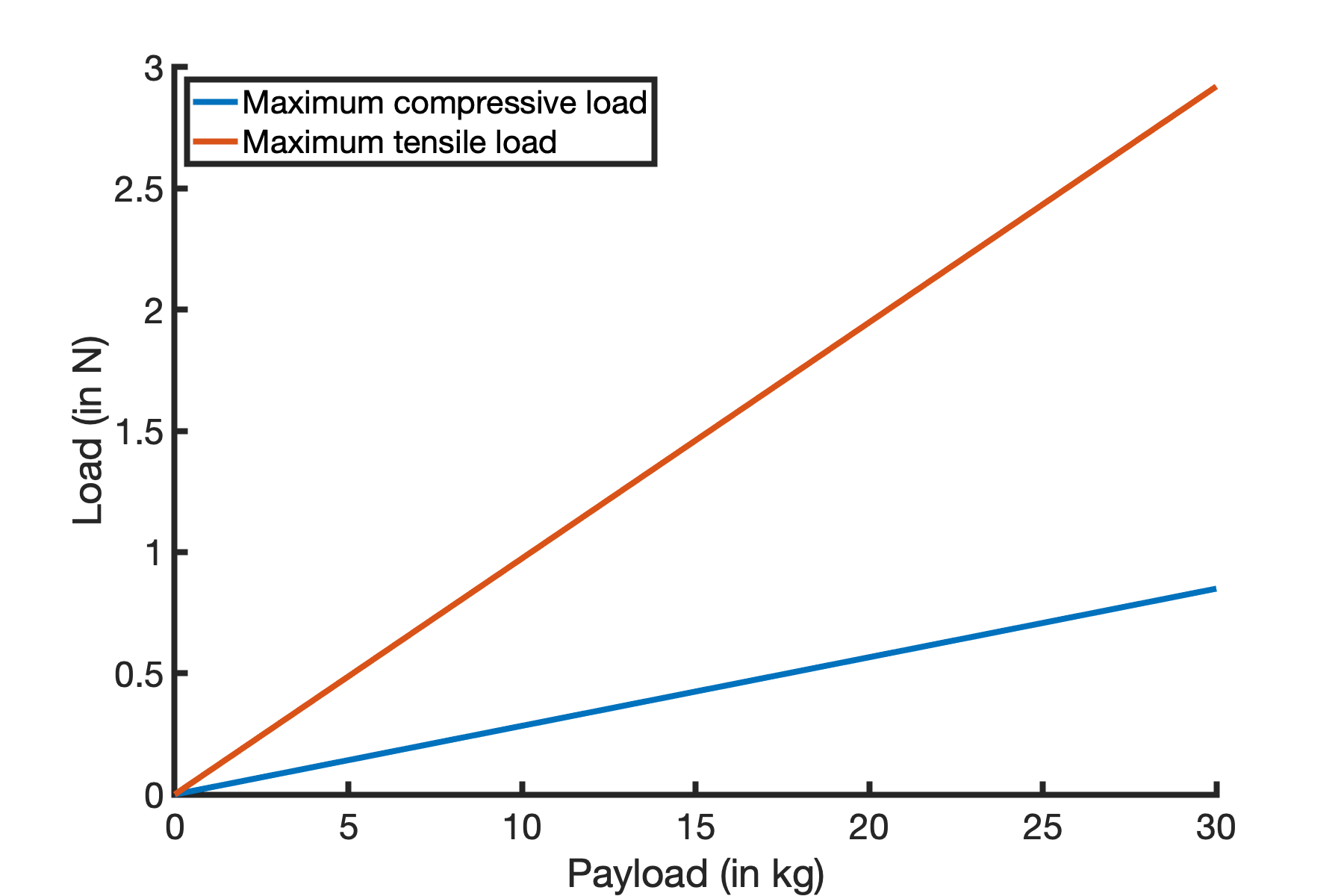}
        \caption{Three-point bottom attachment scenario}
        \label{fig:loads:three_pt}
    \end{subfigure}
    \caption{Maximum compressive and tensile laods in the members of the 2-assembly truss as the payload is increased}
    \label{fig:load:max}
\end{figure}

In the hovering situations, the payload weight is gradually changed from to \SI{0}{\kilo\gram} to \SI{30}{\kilo\gram}, which represents about \SI{2.53}{} the weight of the 2-assembly. Fig.~\ref{fig:load:max} shows the maximum compressive and tensile loads applied to the members of the truss. During all scenarios, the displacements of the nodes remain negligible (under \SI{e-6}{\meter}), such that the structure deformation is minimal. 

\begin{figure}[h!]
    \centering
    \includegraphics[width=0.33\textwidth]{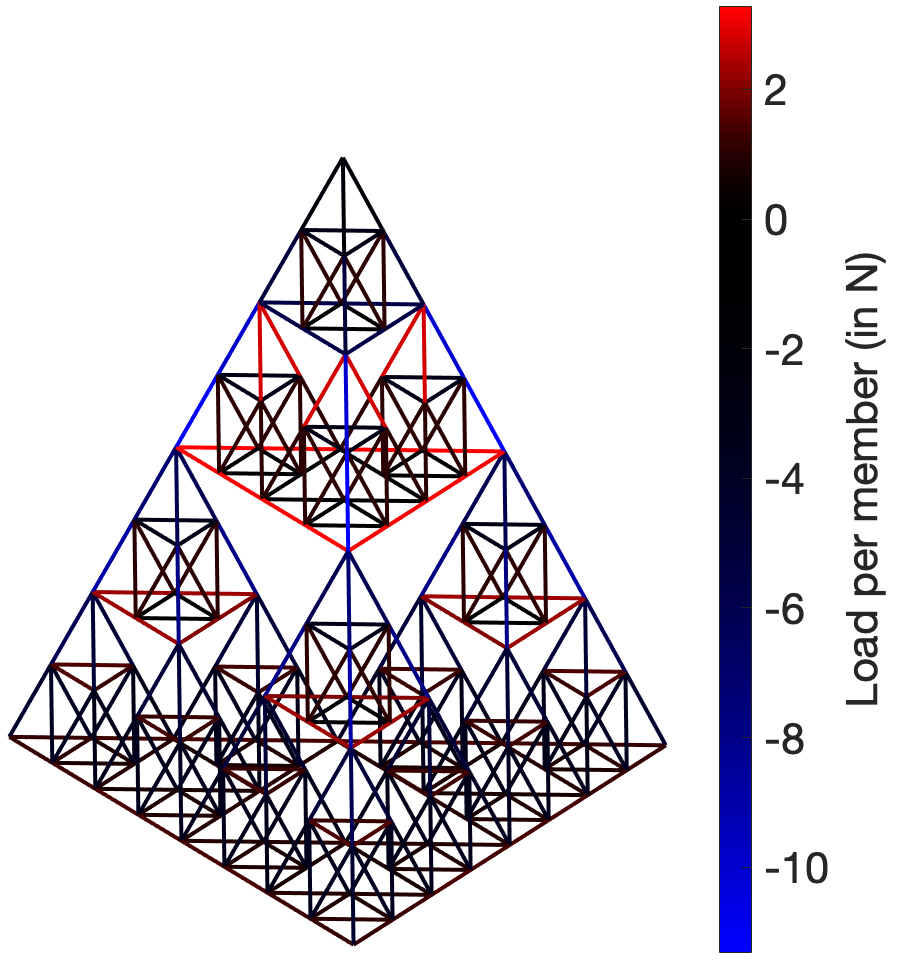}
    \caption{Loads applied on the truss members for the structure at rest. Blue colors indicate compressive forces while red colors indicate tensile forces.}
    \label{fig:loads:at_rest_truss}
\end{figure}

\begin{figure}[h!]
    \centering
    \includegraphics[width=0.33\textwidth]{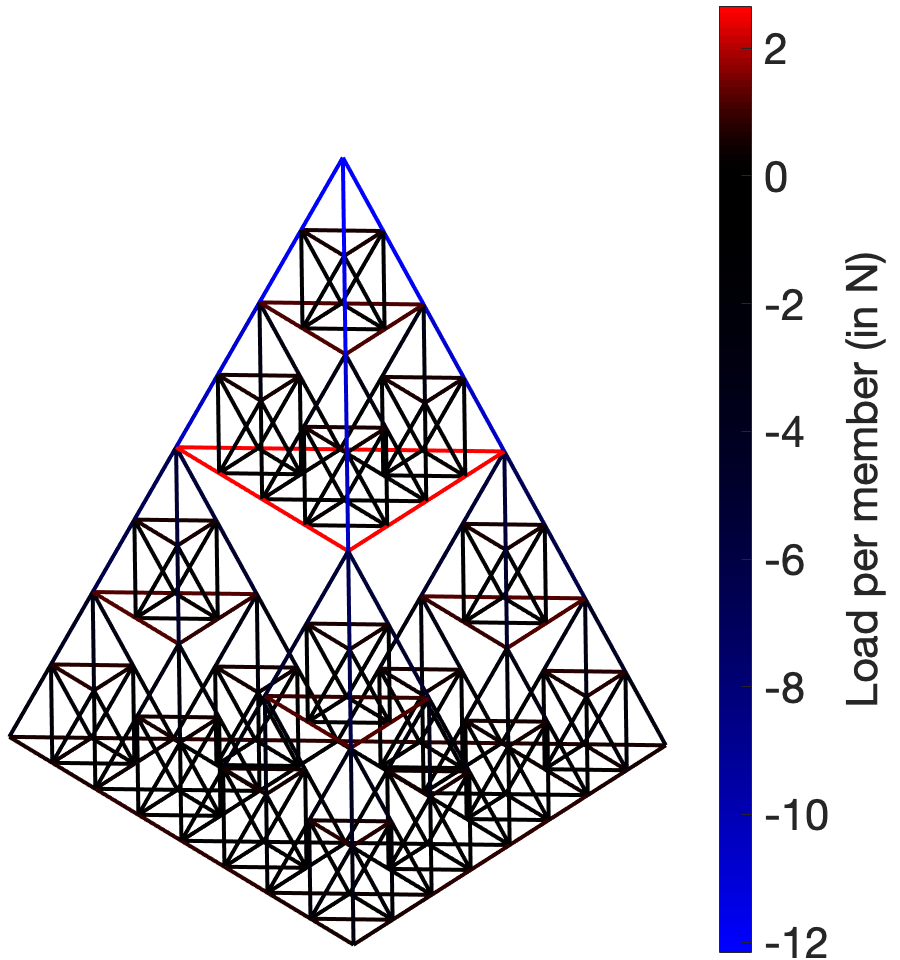}
    \caption{Loads applied on the truss members for the hovering of the structure in the single point top attachment scenario. Payload is equal to \SI{30}{\kilo\gram}, or about \SI{2.53}{} times the assembly mass. Blue colors indicate compressive forces while red colors indicate tensile forces.}
    \label{fig:loads:single_pt_truss}
\end{figure}

\begin{figure}[h!]
    \centering
    \includegraphics[width=0.33\textwidth]{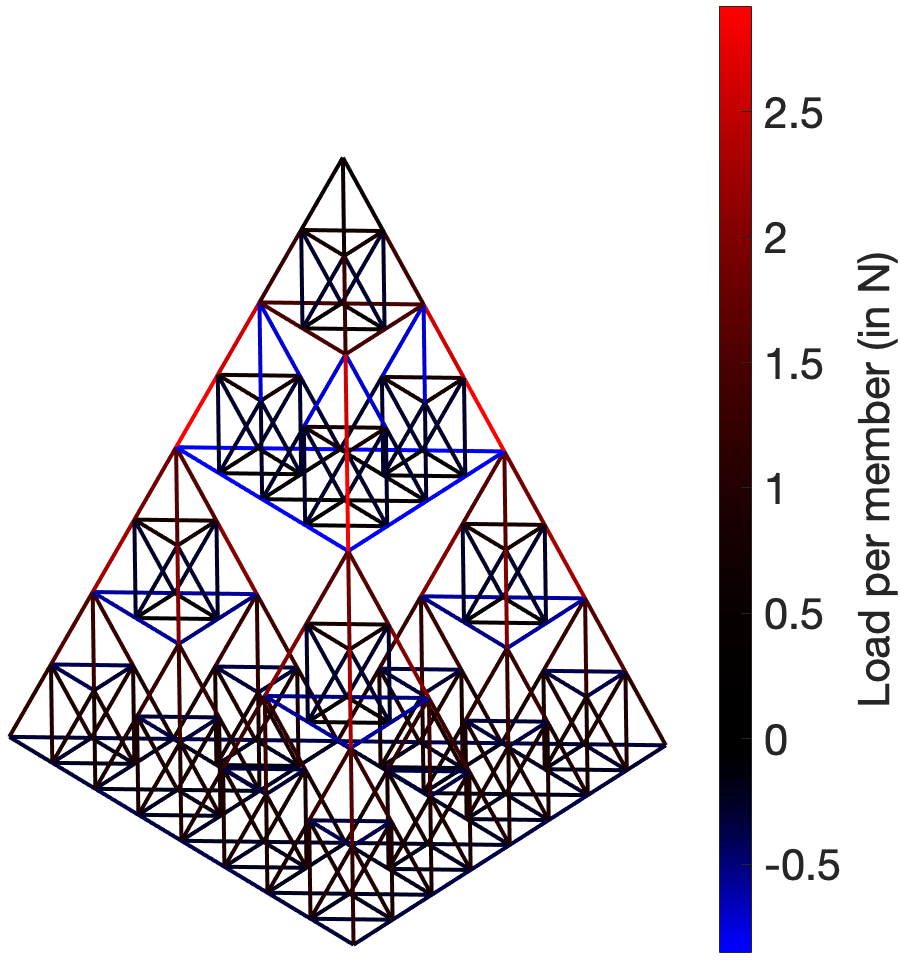}
    \caption{Loads applied on the truss members for the hovering of the structure in the three-point bottom attachment scenario. Payload is equal to \SI{30}{\kilo\gram}, or about \SI{2.53}{} times the assembly mass. Blue colors indicate compressive forces while red colors indicate tensile forces.}
    \label{fig:loads:three_pt_truss}
\end{figure}

\Cref{fig:loads:at_rest_truss,fig:loads:single_pt_truss,fig:loads:three_pt_truss} show the trusses and the loads applied on its members in the three different scenarios. For the two payload lifting scenarios, the payload is set to \SI{30}{\kilo\gram}. We can observe that the top members are subject to higher loads than the other members. However, they still remain significantly under their critical buckling load, equal to \SI{659}{N} when computed with a length factor of \SI{2}{}.


\section{Modeling and control of the Tetracopter}\label{sec:tetra_model}

\subsection{Nonlinear dynamics}

\begin{figure}[h]
    \centering
    \includegraphics[width=0.45\textwidth]{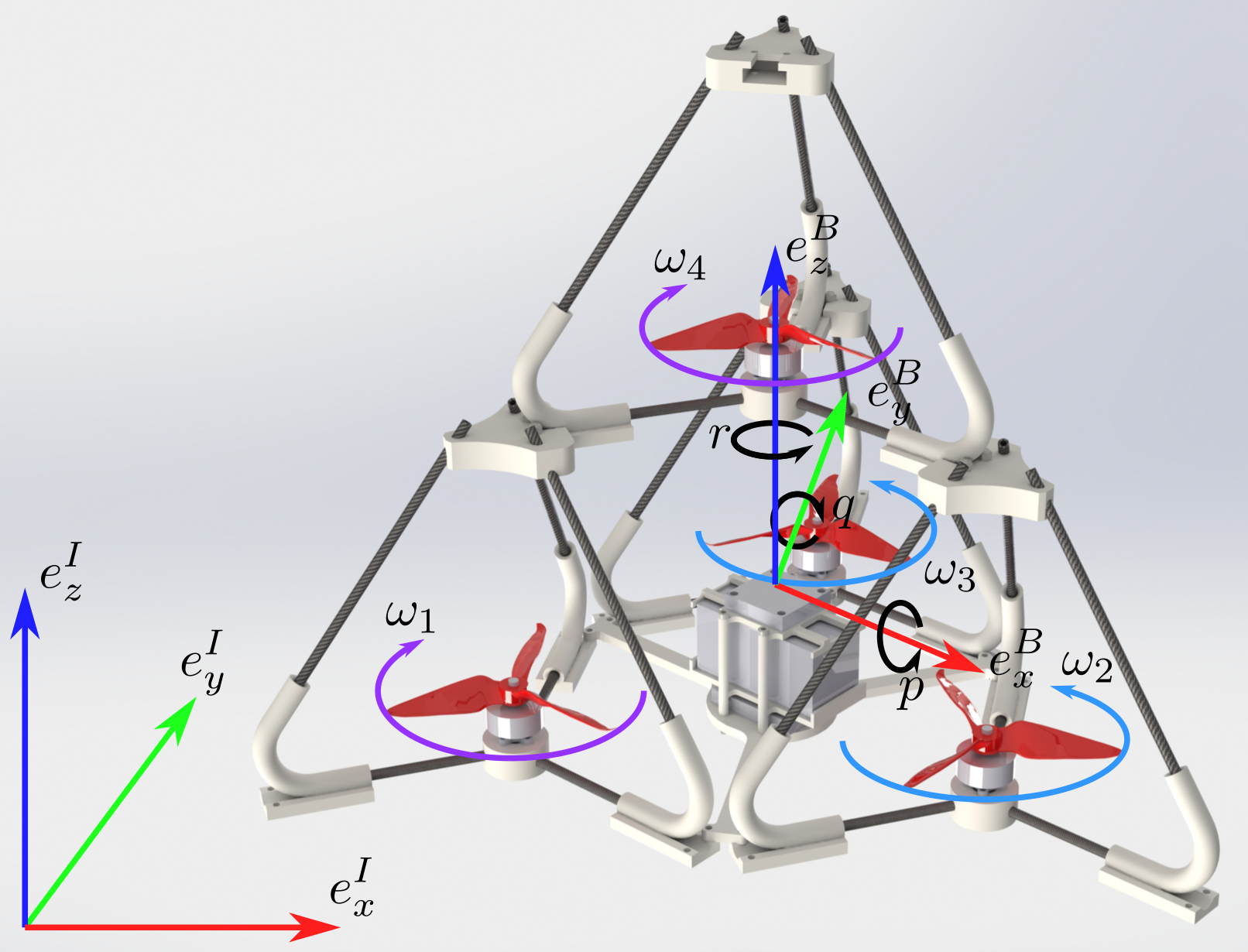}
    \caption{Inertial and body-fixed frames of the Tetracopter.}
    \label{fig:frames}
\end{figure}

The Tetracopter is represented as a rigid body and its dynamics are derived with the Newton-Euler equations.
The derivation of the nonlinear dynamic equations does not differ much from the derivations found in other papers that model flat quadcopters~\cite{luukkonen2011modelling, rinaldi2013linear, chovancova2014mathematical, wang2016dynamics}.
However, the particular placement of the rotors implies a different expression of the torque induced by the differential thrust in \eqref{eq:quad_thrust_torque}.

We consider an inertial reference frame and a body-fixed frame as shown on Fig.~\ref{fig:frames}. The position and orientation of the body-fixed frame in the inertial frame is given by the translation vector $\bm{\xi}$ and the Euler angles $\bm{\eta}$ defined by \eqref{eq:euler_translation} and \eqref{eq:euler_angles}.    

\begin{equation}\label{eq:euler_translation}
    \bm{\xi} = \begin{bmatrix}
        x \\
        y \\
        z \\
    \end{bmatrix}
\end{equation}

\begin{equation}\label{eq:euler_angles}
    \bm{\eta} = \begin{bmatrix}
        \phi \\
        \theta \\
        \psi \\
    \end{bmatrix}
\end{equation}

The rotation matrix from the inertial frame to the body-fixed frame is given by~\eqref{eq:rot_mat}, where the sines and cosines are abbreviated.

\begin{equation}\label{eq:rot_mat}
    R = \begin{bmatrix}
        c_\psi c_\theta & c_\psi s_\theta s_\phi - s_\psi c_\phi & c_\psi s_\theta c_\phi + s_\psi s_\phi \\
        s_\psi c_\theta & s_\psi s_\theta s_\phi + c_\psi c_\phi & s_\psi s_\theta c_\phi - c_\psi s_\phi \\
        -s_\theta & c_\theta s_\phi & c_\theta c_\phi \\
    \end{bmatrix}
\end{equation}

The body-fixed frame has linear velocity $\bm{V^B} = [u, v, w]$ and angular velocity $\bm{\Omega} = [p, q, r]$.

The transformation matrix $S$ is used to obtain the angular velocities in the inertial frame from the angular velocities in the body-fixed frame.
\begin{equation}\label{eq:ang_vel_trans}
    \bm{\Omega} = S \bm{\dot{\eta}} =
    \begin{bmatrix}
        1 & 0 & -s_\theta \\
        0 & c_\phi & c_\theta s_\phi \\
        0 & -s_\phi & c_\theta c_\phi
    \end{bmatrix}
    \begin{bmatrix}
        \dot{\phi} \\
        \dot{\theta} \\
        \dot{\psi} \\
    \end{bmatrix}
\end{equation}

The forces acting on the Tetracopter in the body-fixed frame are gravity, the thrust of the four rotors, and the air drag \cite{chovancova2014mathematical}. Their sum is equal to the centrifugal force and the derivative of the linear momentum in the body-fixed frame:
\begin{equation}\label{eq:quad_forces}
    m\bm{\dot{V}^B} + \bm{\Omega} \times (m \bm{V^B}) = mR^TG + T -
    \begin{bmatrix}
        k_x u^2 \\ k_y v^2 \\ k_z w^2
    \end{bmatrix}.
\end{equation}
The scalars $k_x$, $k_y$, and $k_z$ are drag coefficients.

The thrust produced by the rotors can be written
\begin{equation}
    T =
    k_T\begin{bmatrix}
        0 \\ 0 \\ \omega_1^2 + \omega_2^2 + \omega_3^2 + \omega_4^2
    \end{bmatrix}
\end{equation}
where $k_T$ is a coefficient that depends on the ambient air density and the rotor blades' characteristics and $\omega_j$ is the angular velocity of rotor $j$.

The angular velocity in the body-fixed frame $\bm{\Omega}$ is given by the Euler's equation for a rigid body
\begin{equation}\label{eq:quad_torques}
    \bm{M} = I_q\dot{\bm{\Omega}} + \bm{\Omega}\times(I_q\bm{\Omega})
\end{equation}
where $I_q$ is the inertia tensor of the Tetracopter and $\bm{M}$ the applied torques.

The applied torques include:
\begin{itemize}
    \item $\bm{M}_j$ for $j\in{1,2,3,4}$, the counteracting torques induced on the four stators by spinning the rotor;
    \item $\bm{M^T}$, the torques induced by the differential thrust of the rotors;
    \item $\bm{M^D}$, the drag torque induced by the rotation of the rotorcraft around $\bm{\Omega}$.
\end{itemize} 

For a rotor $j$, $\bm{M}_j$ can be determined with Euler's equation applied to the rotor in the body-fixed frame
\begin{equation}\label{eq:rotor_torques}
    -\bm{M}_j + \bm{M^d}_j + \bm{M^f}_j = I_r \dot{\bm{\omega}_j} + I_r\bm{\Omega} \times \bm{\omega}_j
\end{equation}
where we use:
\begin{itemize}
    \item $\bm{M^d}_j$, the drag torque induced by the rotation of the rotor;
    \item $\bm{M^f}_j$, the friction torque of the rotor with with the stators;
    \item $I_r$, the inertia of a rotor around its rotation axis.
\end{itemize}

Since rotors \si{1} and \si{3} rotate counter-clockwise and rotors \si{2} and \si{4} rotate clockwise,
\begin{equation}\label{eq:rotor_rot}
    \bm{\omega}_j = (-1)^{j+1}\omega_j \mathbf{e^B_z},
\end{equation}
\begin{equation}\label{eq:rotor_drag}
    \bm{M^d}_j = (-1)^j k_D \omega_j^2 \mathbf{e^B_z},
\end{equation}
and
\begin{equation}\label{eq:rotor_friction}
    \bm{M^f}_j = (-1)^j k_F \omega_j \mathbf{e^B_z}
\end{equation}
where $k_D$ and $k_F$ are respectively a drag and a friction constant \cite{rinaldi2013linear, chovancova2014mathematical}.

Combining \cref{eq:rotor_torques,eq:rotor_rot,eq:rotor_drag,eq:rotor_friction} gives
\begin{equation}
    \bm{M}_j 
    =  (-1)^j 
    \begin{bmatrix}
        I_r \omega_j q \\
        - I_r \omega_j p \\
        I_r \dot{\omega_j} + k_D \omega_j^2 + k_F \omega_j
    \end{bmatrix}.
\end{equation}

The torque induced by the drag on the rotorcraft is given by \eqref{eq:quad_drag}, where $k_p$, $k_q$, and $k_r$ are drag coefficients \cite{chovancova2014mathematical}.
\begin{equation}\label{eq:quad_drag}
    \bm{M^D} = -\begin{bmatrix}
        k_p p^2 \\
        k_q q^2\\
        k_r r^2
    \end{bmatrix}.
\end{equation}

\eqref{eq:quad_thrust_torque} gives the torque induced by the differential thrust of the rotors. $a$ is the length of the side of tetrahedral frame and the assumption is made that the z-axis intersects the base of the tetrahedron formed by the rotorcraft in its center.
\begin{equation}\label{eq:quad_thrust_torque}
    \bm{M^T} = a k_T\begin{bmatrix}
        (\omega_3^2 - \omega_1^2)/4  \\
        ((\omega_1^2 + \omega_3^2)/2 - \omega_2^2)/(2\sqrt{3})  \\
        0
    \end{bmatrix}.
\end{equation}

By replacing $\bm{M}$ by $\sum_{j=1}^{4}\bm{M}_j + \bm{M^T} + \bm{M^D}$ in~\eqref{eq:quad_torques}, we obtain
\begin{equation}\label{eq:quad_rotation}
    \begin{split}
        \dot{\bm{\Omega}} =\ &I_q^{-1}
        \bigg( \sum_{j=1}^4 (-1)^j 
            \begin{bmatrix}
                I_r \omega_j q \\
                - I_r \omega_j p \\
                I_r \dot{\omega}_j + k_D \omega_j^2 + k_F \omega_j
            \end{bmatrix}\\
                                 &-\begin{bmatrix}
                                     k_p p^2 \\
                                     k_q q^2\\
                                     k_r r^2
                                 \end{bmatrix}\\
                                 &+ ak_T\begin{bmatrix}
                                     (\omega_3^2 - \omega_2^2)/4  \\
                                     ((\omega_1^2 + \omega_3^2)/2 - \omega_2^2)/(2\sqrt{3})  \\
                                     0
                                 \end{bmatrix}\\
                                 &- \bm{\Omega}\times(I_q\bm{\Omega})
                             \bigg).
    \end{split}
\end{equation}

\subsection{Linearized dynamics}
To derive a stabilizing control law at hover, we can linearize the nonlinear dynamics around $\theta = 0$, $\bm{\Omega} = [0, 0, 0]$ and $\bm{V^B} = [0, 0, 0]$. $\phi$ and $\psi$ can also be assumed equal to zero without loss of generality. We write
\[
    \omega_j = \omega_0 + \Delta \omega_j
\]
where $\omega_0 = \sqrt{mg/(4k_t)}$ so that $T + mG = 0$.

Linearizing~\eqref{eq:ang_vel_trans} gives
\begin{equation}\label{eq:ang_vel_lin}
    \begin{bmatrix}
        \dot{\phi} \\ \dot{\theta} \\ \dot{\psi} \\
    \end{bmatrix}
    =
    \begin{bmatrix}
        p \\ q \\ r
    \end{bmatrix}
\end{equation}
while we also have
\begin{equation}\label{eq:lin_vel_lin}
    \begin{bmatrix}
        \dot{x} \\ \dot{y} \\ \dot{z} \\
    \end{bmatrix}
    =
    \begin{bmatrix}
        u \\ v \\ w
    \end{bmatrix}.
\end{equation}

\eqref{eq:quad_forces} becomes
\begin{equation}\label{eq:forces_lin}
    \begin{bmatrix}
        \dot{u} \\ \dot{v} \\ \dot{w}
    \end{bmatrix}
    =
    \begin{bmatrix}
        g\theta \\
        -g\phi \\
        \frac{2k_T \omega_0}m \sum_{k=1}^4 \Delta \omega_j
    \end{bmatrix}
\end{equation}

and \eqref{eq:quad_rotation} becomes
\begin{equation}\label{eq:torques_lin}
    \begin{split}
        \begin{bmatrix}
            \dot{p} \\ \dot{q} \\ \dot{r}
        \end{bmatrix}
        =\ &I_q^{-1}
        \bigg(  
            \sum_{j=1}^4 (-1)^j \begin{bmatrix}
                I_r\omega_0 q \\
                I_r\omega_0 p \\
                (2k_D \omega_0 + k_F) \Delta \omega_j
            \end{bmatrix}\\
           &+ ak_T\begin{bmatrix}
               (\omega_0 \Delta \omega_3 - \omega_0 \Delta \omega_1)/2  \\
               ((\omega_0 \Delta \omega_1 + \omega_0 \Delta \omega_3)/2 - \omega_0 \Delta \omega_2)/\sqrt{3}  \\
               0
           \end{bmatrix}
       \bigg)
    \end{split}
\end{equation}
where the term $I_r \dot{\Delta \omega_j}$ has been neglected.

\Cref{eq:ang_vel_lin,eq:lin_vel_lin,eq:forces_lin,eq:torques_lin} can be combined together and rewritten as a linear control system whose state is
\[
    [x, y, z, \phi, \theta, \psi, u, v, w, p, q, r]^T
\]
and controls are
\[
    [\Delta \omega_1, \Delta \omega_2, \Delta \omega_3, \Delta \omega_4]^T.
\]

\subsection{Control of the Tetracopter}

The linearized equation of motion can be used to derive a stabilizing feedback control law around the equilibrium point corresponding to hovering. A study of the linear control system obtained previously could be used to derive an optimal control law given a cost function to minimize. However, since the dynamics of the Tetracopter are actually very similar to the ones of a regular quadcopter, a standard quadcopter firmware with a PID controller implementation is used to control the prototype. The PID controller only performs a stabilization of the attitude based on the angular rates. The gains were initialized with the knowledge of the rotor placements and their magnitude for each of the three axes was fine-tuned by performing flight tests.
	
A detailed analysis around different equilibrium points could be performed to derive a more sophisticated and robust control law stabilizing the rotorcraft around different attitudes. Future versions of the Tetracopter will include such an analysis and simulation work to design the control policy.

\subsection{Efficiency Analysis}

\begin{figure}[h]
    \centering
    \begin{subfigure}[b]{0.22\textwidth}
        \includegraphics[width=\textwidth]{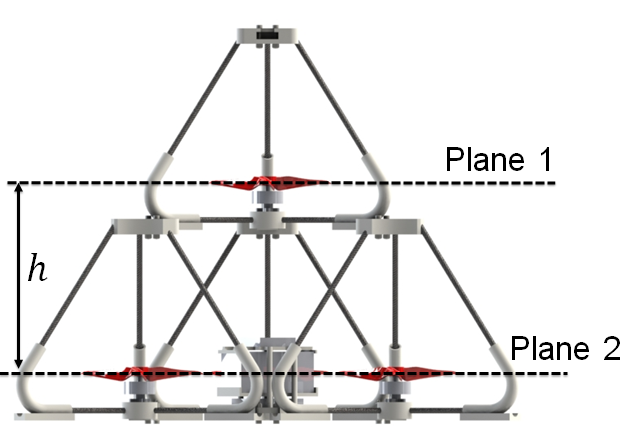}
        \caption{Side view of the Tetracopter with the distance between rotor planes labeled.}
        \label{fig:mu}
    \end{subfigure}
    \begin{subfigure}[b]{0.22\textwidth}
        \includegraphics[width=\textwidth]{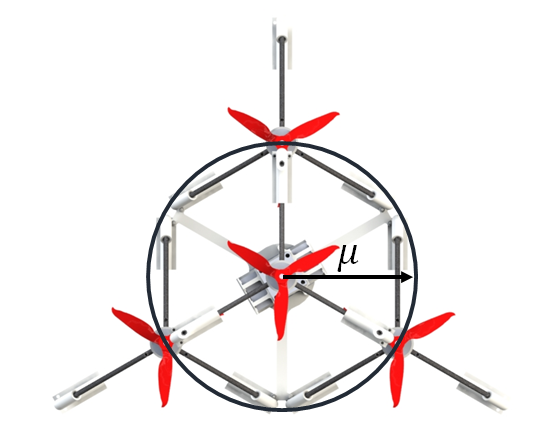}
        \caption{Top view cut section of the Tetracopter with the distance from the center rotor to the outside rotors labeled.}
        \label{fig:h}
    \end{subfigure}
\end{figure}

As the number of elementary modules forming the Tetrahedron Fractal Assembly increases, the concern of wake interactions becomes more apparent. The two parameters changing the impact of the wake interactions are the distance between the rotational axis of the rotors, $\mu$, and the distance between the plane of the rotors, $h$, as labeled in Fig.~\ref{fig:h} and Fig.~\ref{fig:mu} respectively. 
Otsuka and Nagatani experimentally showed that the change in distance between the planes of the rotors has a negligible impact on the thrust generated by the rotors given a constant input voltage to the motors~\cite{Thrustloss}. Otsuka and Nagatani also proved that the distance between the axis of the rotors, $\mu$, must be at least the diameter of the rotor, thus not creating an overlap between the rotors on different planes, for the wake interactions to be negligible. These findings influenced the design of the Fractal Tetrahedron Assembly and the Tetracopter prototype. The rotor configuration of the Tetracopter is optimized for thrust generation, and in theory is relatively identical to a two dimensional rotorcraft in regard to efficiency. The effect of the frame of the single-propeller module on the performance of the enclosed rotor was analyzed using a computation fluid dynamics(CFD) software. The results of the analysis suggest a thrust reduction of 44\% when compared to the same rotor in free flow.  Viewing Fig.~\ref{fig:h}, the rotor coincident with Plane 1 would be the only rotor to experience a 44\% reduction in thrust. From the CFD results, it is shown that the lower rotors on Plane 2 of Fig.~\ref{fig:h} would experience a thrust reduction of approximately 62\% due to the flow obstruction caused by the Tetracopter's frame. 

\section{Tetracopter prototype description and testing}\label{sec:prototype}

 A prototype of the Tetracopter was created to prove that the vehicle would fly in the given configuration. The goal of the prototype was to hover out of ground effect without any noticeable instabilities that could not be repaired through tuning the PID gains on the flight controller.

\subsection{Frame}

\begin{figure}[h]
    \centering
    \begin{tikzpicture}
        \begin{scope}
            \node[anchor=south west, inner sep=0] (image) at (0,0) {\includegraphics[width=0.45\textwidth]{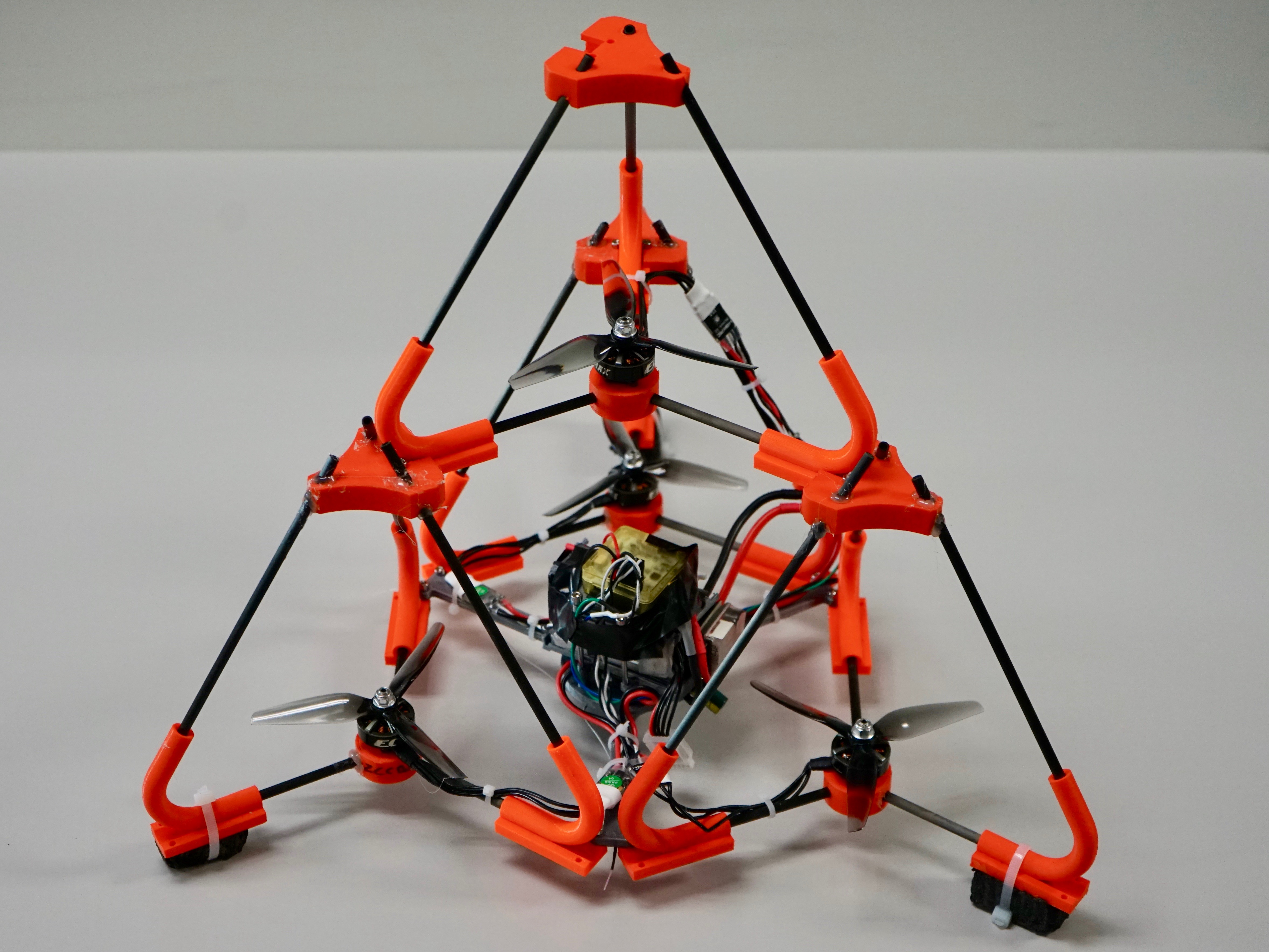}};
            \node[anchor=west, align=center] (feet_text) at (6, 3) {Attachment\\ foot};
            \node[anchor=west, align=center] (top_text) at (5.2, 5.2) {Top attachment\\ piece};
            \draw [-latex, thick] (feet_text) to (6.8, 0.7);
            \draw [-latex, thick] (top_text) to (4.2, 5.7);
        \end{scope}
    \end{tikzpicture}
    \caption{The single-propeller module are attached using these two 3D printed parts}
    \label{fig:attach}
\end{figure}

The frame of the single-propeller submodule takes the shape of a regular tetrahedron. The structure of the single-propeller submodule was fabricated using several 3D printed parts and \SI{5}{\milli\meter}-wide carbon fiber tubes. The design and materials used to build the prototype were meant to minimize blocking the inflow of the rotor as well as reduce the occupied space directly underneath each rotor. The single-propeller submodules are attached by sliding the feet of the aircraft into the top connecting piece. The pieces are then bolted together, as seen in Fig.~\ref{fig:attach}, to form the Tetracopter. 
In the center of the Tetracopter, a 3D printed plate holds the battery, power distribution board, and flight controller. The electronics of the aircraft are positioned close to the center of gravity of the rotorcraft for control purposes.

\subsection{Hardware}

The electronics used are commonly used in conventional quadrotors. The Tetracopter uses four HQProp Ethix S5 5x4x3 rotors and Emax Eco 2306 2400kv motors. The CC3D flight controller along with four HAKRC BLHeli-32 Bit 35A 2-5s electronic speed controllers are used to control the motors.

\subsection{Parameter identification}

\begin{figure}[h]
\centering
\includegraphics[width=7.5cm]{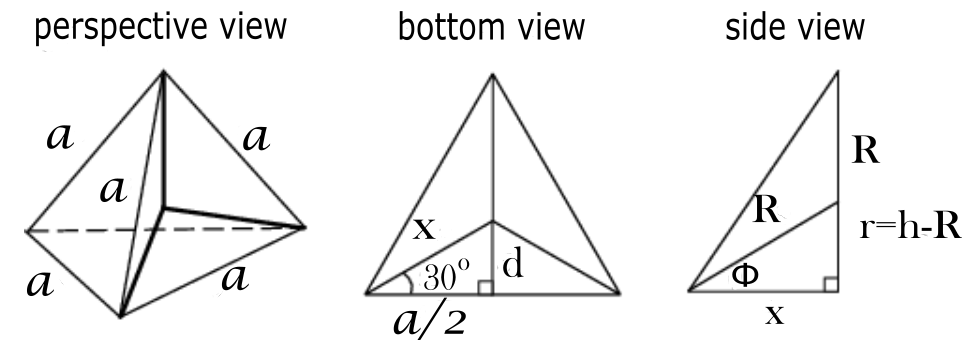}
\caption{Dimensions of a regular tetrahedron~\cite{jackson_weisstein}.}
\label{fig:tet_dim}
\end{figure}

To fully characterize the proposed aircraft a number of parameters must be defined starting with the shape of the frame. As stated before, the frame of the single-propeller module is in the shape of a tetrahedron as seen in Fig.~\ref{fig:tet_dim}. 
Using the variables in Fig.~\ref{fig:tet_dim}, the length of a side of the single-propeller submodule $a$ is equal to \SI{244.55}{\milli\meter}. Given $a$ the remaining variables seen in Fig.~\ref{fig:tet_dim}, can be defined using \cref{eq:geo_1,eq:geo_2,eq:geo_3,eq:geo_4,eq:geo_5}~\cite{jackson_weisstein}.
\begin{equation}
    x=\frac{1}{3}\sqrt{3}a
    \label{eq:geo_1}
\end{equation}
\begin{equation}
    d=\frac{1}{6}\sqrt{3}a
    \label{eq:geo_2}
\end{equation}
\begin{equation}
    h=\frac{1}{3}\sqrt{6}a
    \label{eq:geo_3}
\end{equation}
\begin{equation}
    R=\frac{1}{4}\sqrt{6}a
    \label{eq:geo_4}
\end{equation}
\begin{equation}
    \phi=\tan^{-1}(\frac{r}{x})
    \label{eq:geo_5}
\end{equation}

The distance from the axis of the center rotor to the outer rotors, labeled as $\mu$ in Fig.~\ref{fig:mu}, is approximately \SI{130.4}{\milli\meter} while the distance between the planes labeled in Fig.~\ref{fig:h} as $h$ is \SI{168.8}{\milli\meter}.

The total mass of the vehicle is approximately \SI{740}{\gram}. Given the degradation in thrust due to the frame of the Tetrahedral rotor-craft, at 75\% motor throttle the estimated thrust to weight ratio is 1.94.

\section{Motor fault tolerance}\label{sec:fault_tol}

\begin{figure}[h!]
    \centering
    \subcaptionbox{16 Submodule FTA \label{fig:16RFTA}}
    {\includegraphics[width=4cm]{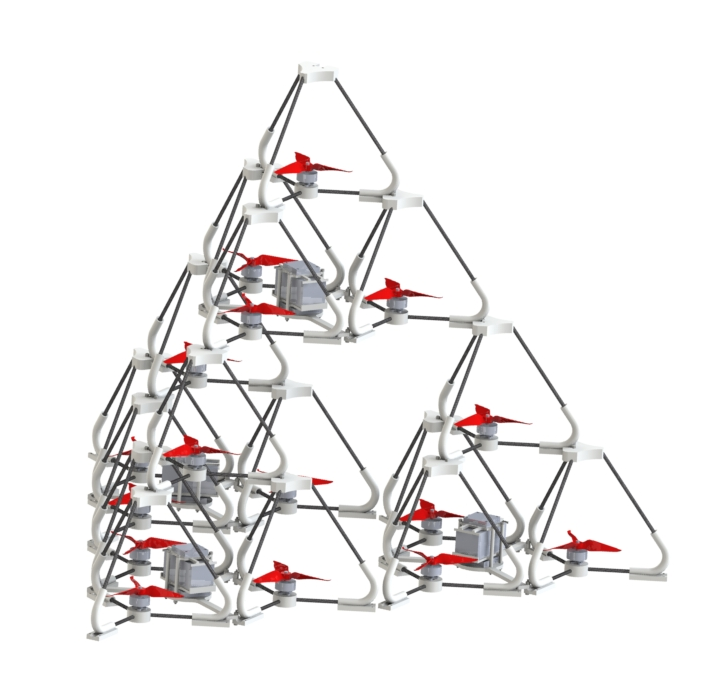}}
    \centering
    \subcaptionbox{16 Submodule FTA Rotor Configuration \label{fig:RotorConfig1}}
    {\includegraphics[width=4cm]{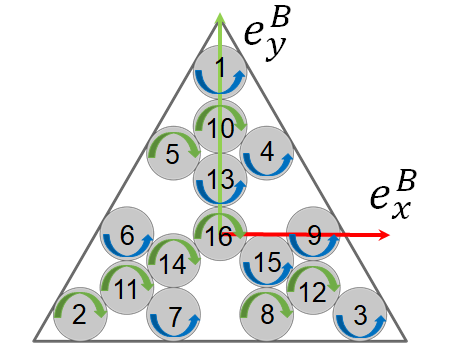}}
\end{figure}

\newcommand\scalemath[2]{\scalebox{#1}{\mbox{\ensuremath{\displaystyle #2}}}}
The rotor configuration of a FTA allows the vehicle to tolerate motor failures. Consider a 16 rotor FTA as shown in Fig.~\ref{fig:16RFTA} with a rotor configuration, shown from a top view, as seen in Fig.~\ref{fig:RotorConfig1}. This rotor configuration is the result of assembling four Tetracopters. For this analysis it is assumed that each rotor can be controlled independently.  

Consider the FTA, shown in Fig.~\ref{fig:RotorConfig1}, in hover, meaning the only forces effecting the FTA are those created by the rotors and gravity. For the FTA to maintain hover in the scenario of a motor failure the overall thrust must be equal to the weight of the vehicle and the roll, pitch, and yaw moments of the vehicle, all expressed in \cref{eq:L,eq:M,eq:N,eq:Z}, must equal zero. 
Using a Computer Aided Design software, the mass $m_1$ of the 16 rotor FTA is estimated to \SI{3.1}{\kilo\gram}. Let $r_{i}^{x}$ and $r_{i}^{y}$ represent the distance motor $i$ is from the body fixed axes $e_{y}^{B}$ and $e_{x}^{B}$ respectively. Along with the constraints on the angular velocity of each rotor, \cref{eq:L,eq:M,eq:N,eq:Z} are used to find the minimum amount of motor failures that would cause the vehicle to be inoperable. To simplify \eqref{eq:N}, it is assumed that the inertia matrix of each rotor is identical and that the cross products of the inertia matrix are equal to zero.
\begin{align}
    \begin{split}\label{eq:L}
        L = &\ \omega_{1}^{2}r_{1y}k-r_{2y}k(\omega_{2}^{2}+\omega_{3}^{2}+\omega_{7}^{2}+\omega_{8}^{2}) \\
            &+r_{4y}k(\omega_{4}^{2}+\omega_{5}^{2}) +\omega_{10}^{2}r_{10y}k \\
            &-r_{11y}k(\omega_{11}^{2}+\omega_{12}^{2})+\omega_{13}^{2}r_{13y}k \\
            &-r_{14y}k(\omega_{14}^{2}+\omega_{15}^{2}) \\
        = &\ 0,
    \end{split}\\
    \begin{split}\label{eq:M}
        M = &\ r_{2x}k(\omega_{2}^{2}-\omega_{3}^{2}) \\
            & + r_{7x}k(\omega_{5}^{2}-\omega_{4}^{2}+\omega_{7}^{2}+\omega_{14}^{2}-\omega_{8}^{2}-\omega_{15}^{2}) \\ 
            & + r_{6x}k(\omega_{6}^{2}-\omega_{9}^{2}+\omega_{11}^{2}-\omega_{12}^{2}) \\
        = &\  0,
    \end{split}\\
    \begin{split}\label{eq:N}
        N =&\ b(\omega_1^2 - \omega_2^2 + \omega_3^2 +\omega_4^2 - \omega_5^2 + \omega_6^2 + \omega_7^2 - \omega_8^2 \\
           &+ \omega_9^2 - \omega_{10}^2 - \omega_{11}^2 - \omega_{12}^2 + \omega_{13}^2 - \omega_{14}^2 + \omega_{15}^2 - \omega_{16}^2) \\
        =&\ 0,
    \end{split}\\
    \begin{split}\label{eq:Z}
        Z=&\ \sum_{n=1}^{16} \omega_{i}^{2}k = m_1,
    \end{split}\\
    \begin{split}\label{eq:lift}
        k=&\ \frac{1}{2} C_{L}r_{prop}^{3} \rho A,
    \end{split}\\
    \begin{split}\label{eq:drag}
    b=&\ \frac{1}{2}C_{D}r_{prop}^{3}\rho A.
    \end{split}
\end{align}

The variables $k$ and $b$ come respectively from the lift and drag equations \eqref{eq:lift} and \eqref{eq:drag}. $C_{L}$ and $C_{D}$ are the lift and drag coefficients of the rotor, $r_{prop}$ is the radius of the rotor, $\rho$ is the density of air at standard atmosphere, and $A$ is the disk area of the rotor.

\Cref{eq:L,eq:M,eq:N,eq:Z} can be rewritten as a single equation by introducing the matrix $D\in\mathbb{R}^{4\times16}$, not written for brevity, and the vectors $\bm{B}$ and $\bm{F}$ defined below. Inequality constraints on the rotors angular speeds are also introduced with the lower and upper bounds $l_b$ and $u_b$ that come from the rotor technical limits. \eqref{eq:SysLin} is as a constrained linear least-squares problem and the existence of a solution guarantees the possibility of an equilibrium point of the system after multiple rotor failures. In \cref{eq:SysLin} the equality constraint indicates which rotor has failed.
\begin{equation}\label{eq:SysLin}
    D\cdot \bm{F} = \bm{B} \quad s.t \begin{cases}
        P \cdot \bm{F} = \bm{0}_{16\times1}\\
        lb \leq \bm{F} \leq ub
    \end{cases}
\end{equation}
\begin{multicols}{2}
    \centering
    \begin{equation*}
        \bm{F}=\begin{bmatrix}
            \omega_{1}^{2} \\ \vdots \\ \omega_{16}^{2}
        \end{bmatrix},
    \end{equation*}\break
    \centering
    \begin{equation*}
        \bm{B}=\begin{bmatrix}
            0 \\ 0 \\ 0 \\ m_1
        \end{bmatrix},
    \end{equation*}
\end{multicols}
\begin{equation}\label{eq:failure}
    P=\begin{bmatrix}
        t_{1} & & (0) \\
              &  \ddots &  \\
        (0) & & t_{16}
    \end{bmatrix},
\end{equation}
\begin{equation*}
    C=0_{16,1}.
\end{equation*}
The variables $lb$ and $ub$ in (\ref{eq:SysLin}) represent the upper and lower bound constraints placed on the angular velocity of the rotors. The variables $t_{i}$ in (\ref{eq:failure}) are boolean variables that specify if a motor has failed (i.e motor 1 failure equates to $t_{1}=1$). Iterating through different motor failure combinations using (\ref{eq:SysLin}), it can be shown that the minimum amount of motor failures needed to hinder the vehicle from hoovering is five. This analysis is also heavily dependent on the weight of the FTA. A heavier FTA would have a lower minimum number of motor failures to hinder steady hover. Although this analysis proves the system robustness, there is a hardware issue that will be encountered as the fractal grows. As the fractal grows in number of rotors, the number of motors to control inputs increases. This would require the development of a flight controller network due to current flight controllers having a finite number of outputs. Coupling rotors so that a several of them receive the same input would allow to use fewer controllers but it would change the minimum number of motor failures needed to make the FTA inoperable.

\section{The 3D-Tetracopter elementary module}\label{sec:3D_module}

The final goal of the research that stems from the presented results, is to develop a fractal tetrahedron assembly made of independent, self-sufficient tetrahedron rotorcraft that can assemble in flight. 

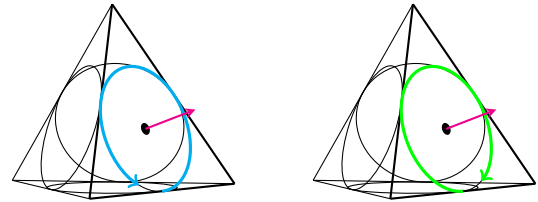
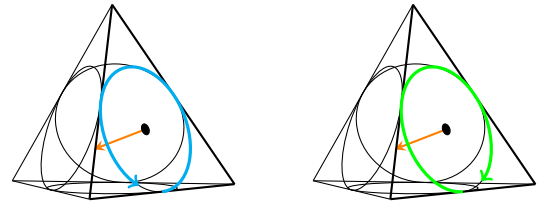
\begin{figure}[h]
   	\begin{subfigure}[b]{0.22\textwidth}
   	\tdplotsetmaincoords{85}{100}
	\begin{tikzpicture}[tdplot_main_coords, scale=3, transform shape, font=\LARGE]
	\pic{tetraprop={0}{0}};
	\end{tikzpicture}
   		\caption{Outward thrust, counterclockwise propeller}
   	\end{subfigure}
   	\begin{subfigure}[b]{0.22\textwidth}
   	\tdplotsetmaincoords{85}{100}
	\begin{tikzpicture}[tdplot_main_coords, scale=3, transform shape, font=\LARGE]
	\pic{tetraprop={0}{1}};
	\end{tikzpicture}
   		\caption{Outward thrust, clockwise propeller}
   	\end{subfigure}
   	\\
   	\begin{subfigure}[b]{0.22\textwidth}
   	\tdplotsetmaincoords{85}{100}
	\begin{tikzpicture}[tdplot_main_coords, scale=3, transform shape, font=\LARGE]
	\pic{tetraprop={1}{0}};
	\end{tikzpicture}
   		\caption{Inward thrust, counterclockwise propeller}
   	\end{subfigure}
   	\begin{subfigure}[b]{0.22\textwidth}
   	\tdplotsetmaincoords{85}{100}
	\begin{tikzpicture}[tdplot_main_coords, scale=3, transform shape, font=\LARGE]
	\pic{tetraprop={1}{1}};
	\end{tikzpicture}
   		\caption{Inward thrust, clockwise propeller}
   	\end{subfigure}
   	\caption{Possible configurations of a propeller for the 4-propeller module}
   	\label{fig:prop_configurations}
\end{figure}

The elementary module presented previously in this paper is already made of four tetrahedra with a propeller attached to each tetrahedron. 
To create a smaller controllable module with a single tetrahedron frame, the possibility of adding three rotors to the single propeller module will be considered. Each rotor can be chosen to spin clockwise or counterclockwise and to produce a thrust directed inward or outward the tetrahedron module. That leaves four possibilities per rotor shown in Fig~\ref{fig:prop_configurations} and a total of $256$ configurations for a 4-propeller tetrahedron module.

\begin{figure}[h]
	\vskip 0pt
   	\begin{subfigure}[b]{0.22\textwidth}
   	\tdplotsetmaincoords{80}{100}
	\begin{tikzpicture}[tdplot_main_coords, scale=3, transform shape, font=\LARGE]
	\pic{tetra={1}{0}{0}{0}};
	\end{tikzpicture}
   	\caption{Configuration A}
   	\label{fig:prop_combinations:A}
   	\end{subfigure}
   	\begin{subfigure}[b]{0.22\textwidth}
   	\tdplotsetmaincoords{260}{100}
	\begin{tikzpicture}[tdplot_main_coords, scale=3, transform shape, font=\LARGE]
	\pic{tetra={0}{1}{1}{1}};
	\end{tikzpicture}
	\caption{Configuration B}
	\label{fig:prop_combinations:B}
   	\end{subfigure}
   	\\
   	\begin{subfigure}[b]{0.48\textwidth}
   	\centering
   	\tdplotsetmaincoords{80}{110}
	\begin{tikzpicture}[tdplot_main_coords, rotate=54, scale=3, transform shape, font=\LARGE]
	\pic{tetra={1}{0}{0}{1}};
	\end{tikzpicture}
	\caption{Configuration C}
	\label{fig:prop_combinations:C}
   	\end{subfigure}
   	\caption{The three stable propeller combinations for the 4-propeller tetrahedron module with counterclockwise propellers.}
   	\label{fig:prop_combinations}
\end{figure}
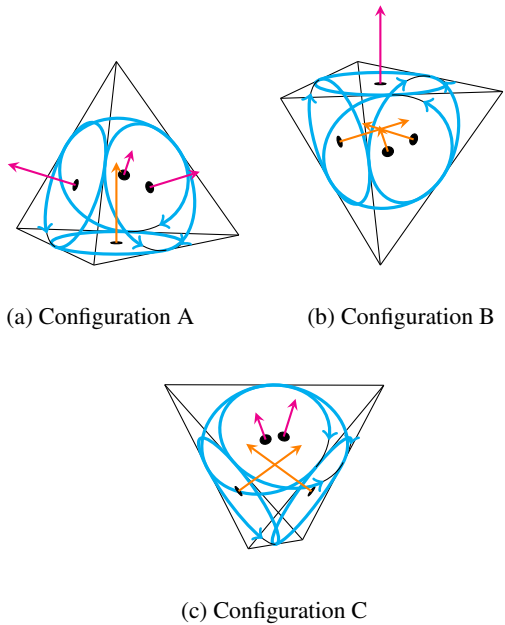

To maintain a steady attitude with the 4-propeller tetrahedron module in a given configuration, the dynamics of the system mush have an equilibrium point. In practice, that means that the propeller configuration must be able to produce a positive thrust output to oppose gravity while generating null total torques.
An analysis of the total moment generated by the four propellers shows that it is null if and only if the propellers spin in the same direction and at the same speed. This constraint eliminates most of the configurations but $32$ of them ($16$ with counterclockwise rotations and $16$ with clockwise rotations).
To reduce the numbers of configurations to study to $16$, we can assume without loss of generality that all propellers spin counterclockwise.
Among these $16$ configurations, the forces equations rules out the two of them where the propellers produce a thrust directed all inward our outward the tetrahedron module, since it wouldn't be able to oppose the gravity force.
The remaining $14$ configurations all have only one equilibrium that corresponds to the orientation of the module in which the total produced thrust when all motors spin at the same speed is opposed to gravity.
By considering the symmetries of the tetrahedron module the number of configurations can be reduced to three. The three possibilities being only determined by the number of propellers producing outward thrust. These three configurations, shown in Fig.~\ref{fig:prop_combinations} are:
\begin{itemize}
	\item Three outward and one inward (Configuration A);
	\item One outward and three inward (Configuration B);
	\item Two outward and two inward (Configuration C).
\end{itemize}

In Configuration A (Fig.~\ref{fig:prop_combinations:A}), the bottom propeller pushes air outward while the three other propellers push air inward so that they all have a positive contribution to the total lift. A geometrical study shows that the lift generated would be twice the thrust of one propeller, assuming that they all produce the same thrust. However, the wake interactions between propellers are likely to affect that number in practice.

Configuration B (Fig.~\ref{fig:prop_combinations:B}) is the opposite of Configuration A and has the same theoretical efficiency. Switching from one of these two configurations to the other would be possible by using symmetrical airfoil propellers and changing their direction of rotation.

Configuration C,  (Fig.~\ref{fig:prop_combinations:C}), has two propellers pushing air outward and two propellers pushing air inward, with the stable hovering attitude corresponding to the tetrahedron module lying on one edge. The total lift in this configuration amounts to $4/\sqrt{3} \approx 2.309$ times the thrust of one propeller. This represents an improvement over the other configurations.
Once again, this represents an estimate of the generated lift from geometrical perspectives only. Airflows and propeller wake interactions can most likely not be neglected for the design and control of such a module. For this reason, future work will include simulations and testings of the three described configurations.

\begin{figure}[h]
\centering
\includegraphics[width=6cm]{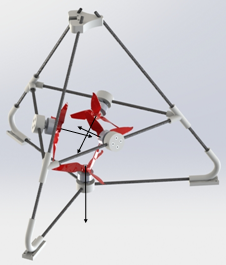}
\caption{A model of a next iteration of the 3D-Tetracopter with three rotors configured to produce force vectors pointing inward while the remaining rotor produces a force vector pointing inward.}
\label{tet_1}
\end{figure}
\begin{figure}[h]
\centering
\includegraphics[width=6cm]{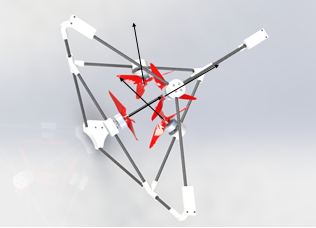}
\caption{A model of a next iteration of the 3D-Tetracopter with two rotors configured to produce force vectors pointing inward while the remaining rotors produce force vectors pointing outward.}
\label{tet_2}
\end{figure}

The control law for the new design of the 4-propeller tetrahedron module will need to be developed and tested to verify that the vehicle can sustain controllable flight. Once this step is completed the hardware and software needed to allow the individual tetrahedron rotorcraft to assemble into a fractal tetrahedron assembly will be researched and experimentally tested. 

\section{Future Work}\label{sec:future_work}

The purpose of this research is to set the foundation for future models of the Tetracopter as well as the Fractal Tetrahedron Assembly.
A prototype of the Fractal Tetrahedron Assembly with multiple Tetracopters will be built and the precise control laws at different level of the assembly will be studied.
Although this research proves that the rotor configuration of the Tetracopter is feasible, there are several parameters of the current design that could increase the efficiency and control of the rotorcraft. One of the main improvements, highlighted by the CFD results, is the aerodynamics of the frame and the relative positioning of the rotors. In the future, the wake interaction between the top propeller of the Tetracopter and the bottom ones will be studied in more details. Reducing the height of the Tetracopter could potentially improve its efficiency, at the cost of a reducuded rigidity when assembled.
Future research will also include more sophisticated hardware for the Tetracopter prototype in order to achieve the goal of in-flight self-assembly.
Other versions of the Tetrahedral rotorcraft using a single tetrahedron with a propeller on each face will also be designed. 
The relationship between the minimum number of motor failures needed to hinder a FTA from hovering and the number of rotors in the FTA is also of interest for future works.

\section{Conclusion}\label{sec:conclusion}

The presented Tetracopter prototype and Fractal Tetrahedon Assembly concept set the foundation for the previously discussed future work. The study of the FTA disk area shows that it can achieve a good theoretical performance as it grows in size while maintaining structural rigidity. A Tetracopter prototype was built and successfully flight tested. Its modular design will allow fast prototyping of bigger assemblies. Another advantage of the FTA shown in this paper is the ability to withstand multiple rotor failures without loss of control. In the future, the assembly and control of the multiple quad-rotorcraft will be studied. Other versions of the Tetrahedral rotorcraft using a single tetrahedron with a propeller on each face will also be designed. 

\printbibliography

\end{document}